%% file: main.tex
\documentclass[runningheads]{llncs}
\usepackage[T1]{fontenc}
\usepackage{graphicx}
\usepackage{booktabs}
\usepackage[misc]{ifsym}
\usepackage[numbers]{natbib}
\usepackage{amsmath}
\usepackage{amssymb}
\usepackage{tabularx}
\usepackage{multirow}

\usepackage{mwe}
\begin{document}

\title{Needles in the Landscape: Semi-Supervised Pseudolabeling for Archaeological Site Discovery under Label Scarcity}

\titlerunning{Semi-Supervised Pseudolabeling for Archaeological Site Discovery}

%N.B.: Author information (both in the \author{} and \authorrunning{} command) should only be present in the Camera-Ready Version of your paper. The version that you initially submit for review, ought to be double-blind. So, when initially submitting your paper, use:
%\author{Author information scrubbed for double-blind reviewing}
\author{Simon Jaxy\inst{1}$^*$ \and 
Anton Theys\inst{2}$^*$ \and
Patrick Willett\inst{3}$^*$ \and
W. Chris Carleton\inst{4} \and
Ralf Vandam\inst{3}$^{\dagger}$ \and
Pieter Libin\inst{1}$^{\dagger}$
}
% You may leave out the orcidID information, if you want to.
% Use \corr to indicate the corresponding author. Note the spacing around the \corr command. Only one author can be the corresponding author.

%N.B.: comment out the \authorrunning{} command for the double-blind version of your paper submitted for review. Later, if your paper is accepted, use the command for the Camera-Ready Version.

\authorrunning{S. Jaxy al.}

% First names are abbreviated in the running head.
% If there is one author, write 'A.L. Benjamin'.
% If there are two authors, write 'A.L. Benjamin and C.C. Broadus Jr.'
% If there are more than two authors, '[...] et al.' is used.

\institute{AI Lab, Department of Computer Science, Vrije Universiteit Brussel \email{\{simon.jaxy\}@vub.be}
\and
Department of Communications
Information Systems and Sensors, Royal Military Academy, Brussels, Belgium 
\and
AMGC (Archaeology, Environmental Changes \& Geo-Chemistry), Vrije Universiteit Brussel
\and
Max Planck Institute of Geoanthropology, Jena, Germany \\
$^*$ Shared first author \\
$^\dagger$ Shared last author
}

\maketitle              % typeset the header of the contribution

\begin{abstract}
\input{sections/00_abstract}


\end{abstract}

\section{Introduction}
\input{sections/01_introduction}

\section{Related Work}
\input{sections/02_background}

\section{Methodology}
\input{sections/04_methodology}

\input{sections/05_experimental_design}

\section{Experimental Results}
\input{sections/06_experimental_results}

\section{Ablation}
\input{sections/07_ablations}

\section{Discussion}\label{sec:discussion}
\input{sections/08_discussion}

\section{Conclusion}
\input{sections/09_conclusion}

\subsubsection*{Acknowledgments}
Scrubbed for double-blind review.

\paragraph*{AI Tools Disclosure.} Generative AI assisted with code development, experiments, and manuscript preparation; all outputs were reviewed by the lead author.

% Commented out for anonymous review
%\begin{credits}
%\subsubsection{\ackname}
%\input{sections/acknowledgement}
%\subsubsection{\discintname}
%
%\end{credits}
%
% ---- Bibliography ----
%
% BibTeX users should specify bibliography style 'splncs04'.
% References will then be sorted and formatted in the correct style.
%
\bibliographystyle{splncs04}
\bibliography{references}
%% Note that this preceding line implies that you store your BibTeX references in a file called 'mybibliography.bib'. If you instead store your references in a file with a different name, for instance 'references.bib', the preceding line should read '\bibliography{references}'. Whatever you do, DO NOT put the file name extension .bib inside the \bibliography command; this will trip up LaTeX compilers. 
%

\newpage
\begin{appendix}

\setcounter{figure}{0}
\renewcommand{\thefigure}{S.\arabic{figure}}

\input{sections/supplementary}

\end{appendix}

\end{document}

%% file: sections/00_abstract.tex
Archaeological predictive modelling estimates where undiscovered sites are likely to occur by combining known locations with environmental and geospatial variables, presenting a positive-unlabeled (PU) learning challenge where confirmed sites are rare and most locations are unlabeled rather than truly negative. To overcome this, we propose asymmetric dual pseudolabeling (DPL), an end-to-end deep learning method that learns from sparse positives directly from multi-band geospatial imagery without hand-crafted feature engineering or assumptions about site absence, and evaluate on two prominent archaeological datasets. On the Sagalassos dataset, evaluated against an independent, held-out field survey, DPL outperforms the LAMAP baseline by 12\% in F1 and 29\% in Recall, while LAMAP maintains advantages in probability ranking. Standard supervised baselines fail catastrophically when negatives are uncertain; positive-only training collapses to predicting everywhere, establishing empirical bounds. On the Cyprus dataset, a pure PU setting without confirmed negatives, SL inverts probability rankings while DPL recovers discrimination. DPL ensembles produce interpretable probability surfaces supporting survey planning, enabling effective site discovery from minimal labeled data.

%% file: sections/01_introduction.tex
Archaeological sites are rare and sparsely distributed across vast, heterogeneous landscapes, leaving behind only limited traces of past human activity. Archaeological Predictive Modeling (APM) seeks to locate these sites by estimating where archaeological sites and artefacts are most likely to occur. The task is inherently difficult: the extreme imbalance between survey areas and known sites, the uncertainty about how many sites exist, and the complexity of landscapes make discovery a formidable challenge.

Traditional methods, like archaeological fieldwalking, remain labor- and time-intensive \cite{banningArchaeologicalSurvey2002}. Statistical approaches, including LAMAP \cite{CARLETON20123371,lirias3727771} and logistic regression \cite{wachtelPredictiveModelingArchaeological2018}, have aided discovery but struggle with high dimensionality and missing absence labels; these challenges often require handcrafted features and strong assumptions \cite{10.1371/journal.pone.0239424,rondeauDoesLocallyAdaptiveModel2022}, motivating end-to-end deep learning approaches that can learn spatial representations directly from raw geospatial data. Data quality issues, such as measurement error and observer bias, further complicate modeling \cite{lirias3727771,rondeauDoesLocallyAdaptiveModel2022}.

Deep learning has advanced in high-dimensional domains, demonstrating an ability to extract complex patterns from multi-modal data. Translating these advances to APM, however, introduces two challenges: \textbf{label sparsity} (sites are <0.004\% of survey areas) undermines standard supervised learning \cite{10.1371/journal.pone.0239424}; and \textbf{positive-only data} (i.e., unlabeled areas are not truly negative) introduces bias. This PU learning challenge parallels ecology (e.g., species distribution with unobserved absences) \cite{phillips2009sample}, clinical medicine (e.g., undiagnosed Brugada syndrome) \cite{10.1093/europace/euad205}, and remote sensing (e.g., sparse ground truth for land use mapping) \cite{Zhao_2023}. APM is thus an instance of a general challenge: learning from incomplete positively biased labels in spatially structured domains.

To address these challenges, we propose a semantic segmentation framework for multi-modal archaeological data under extreme label scarcity. Our approach introduces \textbf{asymmetric pseudolabeling}, which directly mirrors archaeological risk: positive pseudolabels require high confidence (conservative), while negative pseudolabels require only weak evidence (permissive). This enables learning from sparse positives without assuming unlabeled areas are truly negative. Standard supervised baselines (SL) define the optimistic upper bound of performance when confirmed negatives are available, but fail under the survey-completion scenario in a pure PU setting: without confirmed negatives, SL treats all unlabeled background as negative, inverting its probability ranking and scoring background terrain above confirmed sites. Positive-only training (SL\_Pos) defines the conservative lower bound, collapsing to predict sites everywhere regardless of evaluation regime.

Our contributions are as follows:

\begin{enumerate}
\item We frame archaeological predictive modeling as an end-to-end deep learning task, specifically semi-supervised semantic segmentation, under extreme label scarcity and positive-unlabeled uncertainty.
\item We propose an asymmetric pseudolabeling strategy adapted to the archaeological risk of false positives versus false negatives.
\item We demonstrate that standard supervised baselines fail under negative uncertainty, while our approach outperforms the state-of-the-art LAMAP baseline in discovery metrics (F1, Recall).
\item We evaluate on two datasets with complementary validation strategies: Sagalassos, an independent survey validation across seven historical periods; and Cyprus Hellenistic, a pure PU setting without confirmed negatives evaluated under both uniform site-level k-fold (survey completion) and spatial k-fold (geographic generalization). SL\_Pos collapses consistently across both datasets and evaluation regimes. SL fails under survey-completion evaluation in both datasets; under geographic generalization its performance varies widely across spatial folds, confirming that both failure modes are dataset-independent.
\item We release all code and preprocessing instructions publicly; the Cyprus dataset is also released, enabling full reproduction of our results (Sagalassos data is proprietary, but code is shared).
\end{enumerate}

%% file: sections/02_background.tex
\subsection{Deep Learning in Archaeological Predictive Modeling}

Deep learning has been applied to diverse archaeological tasks, including artifact classification, structure detection, and landscape analysis~\cite{landauerArchaeologicalSiteDetection2025}. APM traditionally combines environmental, historical, and remote sensing data with handcrafted features and statistical models such as logistic regression~\cite{wachtelPredictiveModelingArchaeological2018}. While effective in constrained settings, these approaches struggle to scale across heterogeneous landscapes.

Recent work has introduced deep learning into APM via remote sensing imagery~\cite{banasiakSemanticSegmentationUNet2022,bulawkaDeepLearningbasedDetection2024}. Most employ fully supervised training targeting discrete site classes, requiring dense labels or bounding-box annotations; object detection architectures such as YOLOv8~\cite{ultralyticsYOLOv8} are less suited when supervision is limited to sparse point labels and the task requires continuous probability surfaces. Semi-supervised approaches for archaeology exist~\cite{xuSemiSupervisedContrastiveLearning2023}, and pseudolabeling has been applied with a two-cycle retraining scheme~\cite{landauerArchaeologicalSiteDetection2025}; our work differs by generating pseudolabels on-the-fly without retraining. A critical challenge is negative sample selection: random sampling of non-site points introduces significant bias, and KDE-based optimization has been proposed to model settlement influence zones~\cite{zhang2025interpretable}. Our approach instead learns negative pseudolabels directly from model confidence, without spatial assumptions.

\subsection{Pseudolabeling}

Pseudolabeling~\cite{pseudolabels} extends supervision to unlabeled data via model-driven label generation. Dual-branch variants employ dynamic label interpolation~\cite{luo2022scribblesupervisedmedicalimagesegmentation} to stabilize training. We adopt asymmetric thresholding, conservative for positives and permissive for negatives, to learn from sparse positives without assuming background is truly negative.

\subsection{Positive-Unlabeled Learning}

Positive-Unlabeled (PU) learning tackles training with only positive labels while treating the remainder as unlabeled~\cite{elkan2008learning,Bekker_2020}. Standard strategies correct for missing data through risk estimation~\cite{DBLP:journals/corr/KiryoNPS17}; more recent approaches combine PU learning with pseudolabeling or refine predictions through prior distribution comparison~\cite{zhao2022distpupositiveunlabeledlearninglabel}. This challenge recurs across domains: in ecology, Species Distribution Modeling contrasts presence records against background points~\cite{phillips2009sample}; in clinical medicine, rare undiagnosed conditions yield PU data~\cite{10.1093/europace/euad205}; in remote sensing, class prior-free PU methods prevent overfitting when unlabeled data dominates~\cite{Zhao_2023}.

These approaches generally require a known or estimable positive-class prior, and injecting pseudo-absences risks conflating undiscovered with absent~\cite{zhang2025interpretable}. In archaeological contexts, unlabeled locations reflect incomplete surveying rather than true absence, making class-prior estimation infeasible.

%% file: sections/04_methodology.tex
\subsection{Datasets}

\subsubsection{Sagalassos}
The Sagalassos study area in southwestern Turkey covers approximately 3,200 km² (DEM extent), centred on the territory of the ancient city of Sagalassos, with elevation differences exceeding 2,000 m and highly varied topography \cite{vandamMarginalLandscapesHuman2019,c76eb591-d9c7-39f9-8e35-a83c9cadd6a6}. This geologically and culturally complex landscape contains archaeological sites spanning seven chronological periods, documented through decades of systematic archaeological fieldwalking surveys: Late Prehistory (6500--2500 BCE), Iron Age--Archaic (1150--546 BCE), Achaemenid--Hellenistic (546--25 BCE), Roman Imperial (25 BCE--300 CE), Late Antique (300--700 CE), Byzantine (700--1200 CE), and Late Ottoman (1700--1921 CE).

\paragraph{Two independent surveys.} We leverage two distinct archaeological surveys, enabling a critical validation regime. \textbf{Survey 1} (training) derives from over 30 years of systematic archaeological survey \cite{79e093c11d7f4cfa82304dd46566057e}, containing confirmed sites only (PU labels). \textbf{Survey 2} (testing) follows the LAMAP-guided random sampling protocol \cite{lirias3727771}, comprising 84 survey units with 259 total sites across all periods, recording confirmed presences and absences (positive--negative--unlabeled labels). The 13.1\% site overlap (11/84 sites) reflects methodological convergence rather than data leakage. The Sagalassos dataset is proprietary, but all code and preprocessing scripts are released upon publication.

\paragraph{Features.} We extract DEM-derived features (elevation, slope, aspect, curvature, hydrological proximity; 5 channels) from ASTER Global DEM V003 \cite{NASA_ASTER_GDEM_V3}. For periods later than Iron Age--Archaic (1150--546 BCE), we include \textbf{historical infrastructure maps} (distance-to-roads and distance-to-cities), adding 2--3 additional input feature maps depending on the period. All data are resampled to 30~m spatial resolution (EPSG:32636) for pixel-aligned comparison with LAMAP \cite{lirias3727771}.

\paragraph{Training and evaluation radii.} Following LAMAP \cite{lirias3727771}, we dilate point-recorded site coordinates into disk-shaped binary label masks using a training radius of 295~m, matching the spatial uncertainty of field survey records. We define the evaluation radius $r$ as the maximum allowed distance between a predicted positive and the recorded site coordinate for a detection to count. We report results at exact localization ($r=1$~m): a site is detected only if the model assigns a positive prediction to the single pixel nearest to the recorded coordinate, providing the strictest possible spatial criterion.

\subsubsection{Cyprus}
The Cyprus archaeological dataset \cite{crawford2022cyprus} covers the entire island of Cyprus, compiled from multiple archaeological surveys and excavations spanning the island. Here, we focus on the Hellenistic period, selected as a representative period with intermediate site density (i.e., neither the sparsest nor the densest period in known archaeological site count). The Hellenistic period spans two sub-phases: Hellenistic I (312--200 BCE) and Hellenistic II (200 BCE--58 CE). The original extent (4847 $\times$ 8101 pixels, ~39.3M cells at 26.46~m resolution) covers a diverse Mediterranean landscape. We evaluate on a cropped prediction region (2424 $\times$ 2025 pixels, ~4.9M cells) corresponding to the southwest quadrant, providing an independent validation dataset without confirmed negatives.

\paragraph{PU labels and k-fold validation.} Cyprus lacks an independent survey with confirmed negatives. All labels are positive-only (PU setting), requiring spatial k-fold cross-validation. We employ spatial k-fold splitting (K-means++ on cluster centroids) to ensure all sites within the same connected-component cluster remain in the same fold. Spatial separation between folds is enforced by a buffer zone around fold boundaries, reducing spatial autocorrelation leakage between training and validation splits.

\paragraph{Features and data availability.} Cyprus uses DEM-derived features only (elevation, slope, aspect, curvature, water distance) at native 26.46~m resolution without resampling. Since no independent held-out survey with confirmed negatives is available, Cyprus is evaluated via k-fold cross-validation. LAMAP is included as a comparative baseline under the same k-fold protocol. Unlike Sagalassos, where domain experts with direct knowledge of the terrain and site types guided parameter selection, no Cyprus-specific archaeological expertise was available to us; LAMAP parameters were therefore set to principled defaults ($\lambda=1.0$ distance decay, standard-deviation step size) rather than tuned values, and results should be interpreted accordingly. The Cyprus dataset is publicly available \cite{crawford2022cyprus}, and all code and preprocessing scripts are released upon publication, enabling full reproducibility.

\subsection{Model architecture}               
We adopt a UNet architecture \cite{ronneberger2015unetconvolutionalnetworksbiomedical} with an encoder-decoder structure and skip connections for semantic segmentation. Our default backbone is ResNet-18, selected for its favorable trade-off between capacity and computational efficiency under label-scarce conditions. In ablations, we explore smaller (MobileNetV2) and larger (ResNet-50) backbones to assess the impact of model capacity on performance.

The model maps input tiles $\mathbf{x} \in \mathbb{R}^{H \times W \times C}$ (height $H$, width $W$, input channels $C$) to segmentation maps $\hat{\mathbf{y}} \in [0,1]^{H \times W}$ via \begin{equation} \mathcal{D}_{\psi}(\mathcal{E}_\theta(\mathbf{x})) = \hat{\mathbf{y}}, \end{equation} where $\mathcal{E}_\theta$ and $\mathcal{D}_{\psi}$ denote the encoder and decoder, parameterized by neural network weights $\theta$ and $\psi$.

For DPL, we employ a \textbf{dual decoder} architecture: two independent decoders $\mathcal{D}_{\psi_1}$ and $\mathcal{D}_{\psi_2}$ share a common encoder $\mathcal{E}_\theta$ but have independent parameters. This enables diverse predictions for pseudolabel generation while maintaining efficiency through encoder sharing. For SL and SL\_Pos, we use a single decoder.

\subsection{Training Strategies}

\subsubsection{Supervised learning (SL)} 
Let $\{(\mathbf{x}_i, \mathbf{y}_i)\}_{i=1}^{N}$ denote the labeled tiles. The supervised loss is
\begin{equation}
  \mathcal{L}_{\text{SL}} = \frac{1}{N}\sum_{i=1}^{N}
  \ell(\hat{\mathbf{y}}_i, \mathbf{y}_i),
\end{equation}
where $\ell(\cdot, \cdot)$ is binary cross-entropy with positive class weighting to account for label imbalance. SL trains on both positive and negative patches, assuming that background regions are true negatives. This represents an \textbf{optimistic upper bound}: it performs well when confirmed negatives are reliable, but is expected to fail when this assumption is violated.

\subsubsection{Positive-only Supervised Learning (SL\_Pos)}
SL\_Pos trains on positive patches only, addressing the uncertainty about whether background regions are truly negative. Crucially, SL\_Pos differs from SL in its loss computation, where the loss is computed only on positive pixels (i.e., $y = 1$):
\begin{equation}
  \mathcal{L}_{\text{SL\_Pos}} =
  \frac{1}{|\mathcal{P}|}\sum_{(i,j) \in \mathcal{P}}
  \ell(\hat{y}_{ij}, 1),
    \label{eq:sl_pos}
\end{equation} 
where $\mathcal{P} = \{(i,j) : y_{ij} = 1\}$ is the set of positive pixel locations.

This prevents the model from learning to suppress background regions, which may contain undiscovered sites. Without negative supervision, SL\_Pos is expected to predict positive everywhere at test time, making it a \textbf{conservative lower bound}: the limit of what happens when all assumptions about site absence are avoided.

\subsubsection{DPL}

Our dual-branch model produces two independent
predictions:
\begin{equation}
  \hat{\mathbf{y}}^{(1)} =
  \sigma(\mathcal{D}_{\psi_1}(\mathcal{E}_\theta(\mathbf{x}))),
  \quad
  \hat{\mathbf{y}}^{(2)} =
  \sigma(\mathcal{D}_{\psi_2}(\mathcal{E}_\theta(\mathbf{x}))),
\end{equation}
where $\hat{\mathbf{y}}^{(k)} \in [0,1]^{H \times W}$ is the pixel-wise probability map from decoder branch $k$, $\mathcal{E}_\theta$ is the shared encoder, and
$\mathcal{D}_{\psi_1}, \mathcal{D}_{\psi_2}$ are independent decoders.

\paragraph{Asymmetric adaptive thresholds.}
We employ decoupled threshold schedules that evolve during
training via a sigmoid ramp-up function $\rho(t) \in [0,1]$.
The positive threshold increases (more selective) while the
negative threshold decreases (more selective):
\begin{align}
  \tau_{\text{pos}}(t) &= \tau_{\text{pos,start}} +
  \rho(t) \cdot (\tau_{\text{pos,end}} -
  \tau_{\text{pos,start}}) \\
  \tau_{\text{neg}}(t) &= \tau_{\text{neg,start}} -
  \rho(t) \cdot (\tau_{\text{neg,start}} -
  \tau_{\text{neg,end}}).
\end{align}

Threshold schedules and their sensitivity to the starting values are analyzed in the ablation study (Supplementary Section~\ref{supp:ablation}).

\paragraph{Confidence masks via min-prediction.}
  We compute the minimum prediction across branches to identify
  confident regions:
  \begin{align}
  \mathbf{m}_{\text{min}} &= \min(\hat{\mathbf{y}}^{(1)},
  \hat{\mathbf{y}}^{(2)}) \\
  \mathbf{m}_{\text{pos}} &= \mathbf{1}[\mathbf{m}_{\text{min}} >
   \tau_{\text{pos}}] \\
  \mathbf{m}_{\text{neg}} &= \mathbf{1}[\mathbf{m}_{\text{min}} <
   \tau_{\text{neg}}].
  \end{align}
Positive pseudolabels require \textit{consensus} (both
branches confident), while negative pseudolabels require only \textit{dissent} (at least one branch skeptical).

\paragraph{Asymmetric pseudolabel generation.}
  We generate pseudolabels $\tilde{\mathbf{y}}$ via convex
  combination with random coefficient $\beta \sim
  \mathcal{U}(0,1)$:
  \begin{equation}
  \mathbf{w} = \beta \hat{\mathbf{y}}^{(1)} + (1-\beta)
  \hat{\mathbf{y}}^{(2)}.
  \end{equation}
  Crucially, we apply \textbf{asymmetric treatment}: positive
  pseudolabels are temperature-scaled (soft), while negative
  pseudolabels are hard-coded to 0:
  \begin{align}
  \tilde{\mathbf{y}}[\mathbf{m}_{\text{pos}}] &=
  \sigma\!\left(\frac{1}{T}\log\frac{\mathbf{w}[\mathbf{m}_{\text{pos}}]}{1-\mathbf{w}[\mathbf{m}_{\text{pos}}]}\right)
  \quad \text{(soft, $T>1$)} \\
  \tilde{\mathbf{y}}[\mathbf{m}_{\text{neg}}] &= 0 \quad
  \text{(hard)},
\end{align}
where $T>1$ is the pseudolabel temperature, selected via ablation.

\paragraph{Loss function.}
  The total training loss combines four terms:
  \begin{equation}
  \mathcal{L}_{\text{DPL}} = \mathcal{L}_{\text{SL\_Pos}} + \lambda_p
   \mathcal{L}_{\text{pseudo}} + \lambda_c
  \mathcal{L}_{\text{cons}} + \lambda_e \mathcal{L}_{\text{ent}}.
  \label{eq:dpl_full_loss}
\end{equation}

The supervised loss:
\begin{equation}
    \mathcal{L}_{\text{SL\_Pos}} = \frac{1}{2}(\ell(\hat{\mathbf{y}}^{(1)}, \mathbf{y}) +
  \ell(\hat{\mathbf{y}}^{(2)}, \mathbf{y})).
\end{equation}

The pseudolabel loss is defined as
\begin{equation}
\mathcal{L}_{\text{pseudo}} =
  \frac{1}{2}(\ell(\hat{\mathbf{y}}^{(1)}, \tilde{\mathbf{y}}) +
  \ell(\hat{\mathbf{y}}^{(2)}, \tilde{\mathbf{y}})),
\end{equation}
using a focal loss as $\ell$, which down-weights easy, high-confidence background predictions and focuses learning on uncertain pixels (hyperparameters selected via ablation; see Supplementary Section~\ref{supp:ablation}). 

The consistency regularization term enforces agreement between the two prediction branches:
\begin{equation}
    \mathcal{L}_{\text{cons}} = \|\hat{\mathbf{y}}^{(1)} -
  \hat{\mathbf{y}}^{(2)}\|_2^2.
\end{equation}

Finally, entropy regularization is applied on pixels where the two branches disagree:

\begin{equation}
\mathcal{L}_{\text{ent}} =
  \frac{1}{2}(\mathcal{H}(\hat{\mathbf{y}}^{(1)}) +
  \mathcal{H}(\hat{\mathbf{y}}^{(2)})) \odot
  \mathbf{1}[|\hat{\mathbf{y}}^{(1)} - \hat{\mathbf{y}}^{(2)}| >
  0.5],
\end{equation}
where $\mathcal{H}(\cdot)$ denotes binary entropy and $\odot$ element-wise multiplication.

%% file: sections/05_experimental_design.tex
\subsection{Evaluation Protocols}\label{sec:experimental_design}

\subsubsection{Sagalassos: DL vs LAMAP.} We train on Survey 1 and test on Survey 2 (84 survey units with confirmed presences and absences), enabling direct LAMAP comparison using standard metrics (F1, precision, recall, PR-AUC, ROC-AUC). This protocol is applied independently for each of the seven historical periods. We report Late Antique as the primary period (most label-rich, 113 sites); results across all seven periods are reported in Table~\ref{tab:dpl_vs_lamap_r1}.

\subsubsection{Cyprus: Complementary k-fold validation strategies.}
Cyprus lacks an independent survey with confirmed negatives, requiring internal validation. We employ two complementary k-fold strategies to answer distinct questions.

\paragraph{Uniform site-level k-fold (primary).} To test the practical scenario of completing a partially surveyed landscape, we randomly assign 95 site clusters to 5 balanced folds (19 sites per fold). Models train on 76 sites (4 folds) with masked test-fold sites, then predict over the full landscape. Fold-aware evaluation excludes train-fold sites from metrics, preventing data leakage. This matches the survey completion scenario: given known sites, can we locate the remaining ones?

\paragraph{Spatial k-fold with buffer zones (secondary).} To test geographic generalization, we partition the landscape into 5 spatial regions via K-means++ clustering of site centroids, with 4\,km buffer zones creating geographic gaps (0\% patch overlap). Models train on 4 regions and test on the held-out region. Four sites (4.2\%) fall within buffer zones and are excluded. Supplementary Figures~\ref{fig:supp:kfold_uniform} and~\ref{fig:supp:kfold_spatial} illustrate both strategies.

Both strategies use ROC-AUC (positive vs background) and recall at exact localization (within 1\,m of site centroids) as primary metrics. Uniform k-fold serves as our primary result; spatial k-fold provides a geographically conservative complement.

\subsection{LAMAP Baseline}

The Locally Adaptive Multivariate Assemblage Probability (LAMAP) method \cite{CARLETON20123371, lirias3727771} is the state-of-the-art probabilistic baseline for archaeological site prediction. It estimates archaeological potential via joint empirical cumulative distribution functions (JECDFs) over environmental variables, weighted by exponential distance decay from known sites. LAMAP is deterministic and requires no training. It produces probability surfaces over the landscape indicating how likely it is to find a new site at each location; these surfaces have been validated through independent field survey on the Sagalassos study area \cite{lirias3727771}. It serves as a strong DEM-based baseline, though extending it to raw multispectral imagery is non-trivial as it relies on hand-crafted environmental features.

\subsection{Sampling and Training Strategy}

\subsubsection{Hybrid sampling}
To address extreme class imbalance (confirmed sites represent <0.004\% of landscape pixels; training labels are expanded to a 295\,m radius for stable supervision), we employ hybrid sampling: 10\% of training patches are centered on known sites, 90\% randomly sampled from the landscape. Full implementation details (optimizer, learning rate schedule, batch size, DPL hyperparameters) are in Supplementary Section~\ref{supp:implementation}.

\subsubsection{Patch-based training}
Input imagery is divided into $128 \times 128$ tiles sampled via TorchGeo's samplers \cite{Stewart_TorchGeo_Deep_Learning_2024}. For training, we use random batch sampling with hybrid class balancing. For validation and testing, we use systematic grid sampling to generate continuous probability surfaces across the landscape.

%% file: sections/06_experimental_results.tex
\subsection{Sagalassos}

\subsubsection{Metric Results}

Table \ref{tab:phase2_late_antique_r1} shows Late Antique results (the most label-rich period with 113 sites). All deep learning results report mean $\pm$ std over five random seeds (implementation details: Supplementary Section~\ref{supp:implementation}).

\begin{table}[h]
\centering
\caption{Phase 2 Results: Late Antique Period ($r=1$m). All models evaluated on independent Survey 2. Deep learning results are mean $\pm$ std across five seeds. Bold indicates best performance per metric. We mark SL\_Pos positive collapse with $^*$.}
\label{tab:phase2_late_antique_r1}
\begin{tabularx}{\textwidth}{l*{5}{>{\centering\arraybackslash}X}}
\toprule
Model & F1 & Dice & PR-AUC & ROC-AUC & Recall \\
\midrule
LAMAP 
& 0.777 
& 0.777 
& $\mathbf{0.985}$ 
& $\mathbf{0.884}$ 
& 0.643 \\

DPL 
& $\mathbf{0.874 \pm 0.042}$ 
& $\mathbf{0.874 \pm 0.042}$ 
& $0.937 \pm 0.008$ 
& $0.601 \pm 0.015$ 
& $\mathbf{0.831 \pm 0.077}$ \\

SL 
& $0.261 \pm 0.016$ 
& $0.261 \pm 0.016$ 
& $0.950 \pm 0.011$ 
& $0.627 \pm 0.073$ 
& $0.150 \pm 0.011$ \\

SL\_Pos
& $0.955 \pm 0.000$ 
& $0.955 \pm 0.000$ 
& $0.911 \pm 0.011$ 
& $0.521 \pm 0.017$ 
& $^*1.000 \pm 0.000$ \\
\bottomrule
\end{tabularx}
\end{table}

SL misses 85\% of known sites and SL\_Pos collapses to predicting everywhere, confirming the empirical bounds. DPL improves F1 (+12.5\%) and recall (+29.3\%) over LAMAP; LAMAP leads in probability ranking (PR-AUC 0.985 vs.\ 0.937, ROC-AUC 0.884 vs.\ 0.601), reflecting its explicit probabilistic modeling.

\subsubsection{Evaluation Across All Periods}

Table \ref{tab:dpl_vs_lamap_r1} compares DPL and LAMAP across all seven historical periods. DPL outperforms LAMAP in F1 across all periods (gains: 9.6--31.9\%) and Recall (gains: 15--64\%).

\begin{table}[h]
\centering
\caption{DPL vs LAMAP across all periods ($r=1$m).}
\label{tab:dpl_vs_lamap_r1}
\begin{tabularx}{\textwidth}{l*{6}{>{\centering\arraybackslash}X}}
\toprule
& \multicolumn{2}{c}{F1} 
& \multicolumn{2}{c}{PR-AUC} 
& \multicolumn{2}{c}{Recall} \\
\cmidrule(lr){2-3} \cmidrule(lr){4-5} \cmidrule(lr){6-7}
Period 
& LAMAP & DPL 
& LAMAP & DPL 
& LAMAP & DPL \\
\midrule
Iron Age--Archaic   & 0.762 & $\mathbf{0.983}$ & $\mathbf{1.000}$ & 0.988 & 0.615 & $\mathbf{0.976}$ \\
Ottoman            & 0.696 & $\mathbf{0.918}$ & $\mathbf{0.971}$ & 0.925 & 0.552 & $\mathbf{0.887}$ \\
Byzantine          & 0.721 & $\mathbf{0.908}$ & $\mathbf{0.986}$ & 0.896 & 0.571 & $\mathbf{0.938}$ \\
Achaemenid/Hell.   & 0.682 & $\mathbf{0.866}$ & $\mathbf{0.999}$ & 0.946 & 0.518 & $\mathbf{0.807}$ \\
Late Antique       & 0.777 & $\mathbf{0.874}$ & $\mathbf{0.985}$ & 0.937 & 0.643 & $\mathbf{0.831}$ \\
Late Prehistory    & 0.777 & $\mathbf{0.852}$ & 0.979 & 0.906 & 0.651 & $\mathbf{0.812}$ \\
Roman Imperial     & 0.703 & $\mathbf{0.818}$ & $\mathbf{0.978}$ & 0.927 & 0.549 & $\mathbf{0.749}$ \\
\bottomrule
\end{tabularx}
\end{table}

\subsubsection{Qualitative Analysis}

\begin{figure}[h]                                                                                         \centering
  \includegraphics[width=\textwidth]{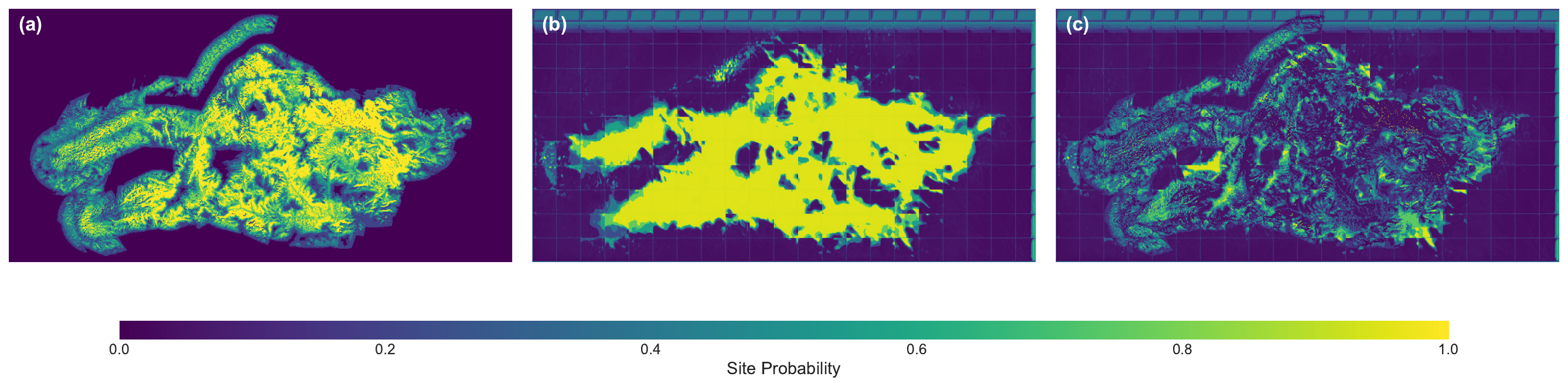}
  \caption{Late Antique probability maps: (a) LAMAP, (b) DPL ensemble mean across 5 seeds,
  (c) absolute difference $|$DPL $-$ LAMAP$|$. All panels share a common viridis colour scale
  (0--1). DPL predicts higher probabilities across more of the landscape, reflecting its higher
  recall at the cost of less conservative probability ranking.}
  \label{fig:phase2_probmaps}
\end{figure}

Figure \ref{fig:phase2_probmaps} compares probability surfaces for Late Antique. LAMAP produces conservative predictions; DPL generates higher-confidence predictions extending beyond known sites. The DPL surface traces the Hellenistic road network, consistent with distance-to-road being the most discriminative feature (Supplementary Section~\ref{supp:feature_separability}). Seed-level variance is low near known sites and higher in intermediate-probability regions (Supplementary Figure~\ref{fig:supp:variance}), suggesting robustness where it matters most.

\subsubsection{Additional Results}
Complementary metric analysis is in Supplementary Section~\ref{supp:radar} and period-wise comparison in Supplementary Section~\ref{supp:period_performances}.

\subsection{Cyprus}\label{sec:phase1}

The Cyprus Hellenistic dataset provides an internal positive-unlabeled validation setting: 167 confirmed Hellenistic site locations form 95 connected-component clusters at the 295\,m training label radius. We evaluate under two complementary k-fold strategies.

\subsubsection{Uniform site-level k-fold (primary)}

Table~\ref{tab:phase1_cyprus_uniform} reports results (implementation: Supplementary Section~\ref{supp:implementation}). SL produces near-zero recall (ROC-AUC $= 0.494 \approx 0.5$) and SL\_Pos collapses (Recall $= 0.999$, ROC-AUC $= 0.497$), establishing the empirical bounds.

DPL achieves ROC-AUC $= 0.510 \pm 0.031$, nominally above both bounds, but with Recall $= 0.239 \pm 0.333$. We disable consistency regularization (it provides no benefit in this pure PU setting; entropy minimization already acts as the primary commitment signal) and use $\tau_{\text{neg,start}}=0.4$, consistent with the Sagalassos-optimal configuration. The high recall variance reveals training instability: in some folds DPL collapses toward SL\_Pos behavior, in others toward near-zero predictions. Cyprus presents a more challenging landscape for pseudo-label bootstrapping. A feature separability analysis (Supplementary Section~\ref{supp:feature_separability}) reveals a counterintuitive pattern: Cyprus DEM and slope are actually more separable than their Sagalassos counterparts (DEM AUC $= 0.799$ vs.\ $0.540$; slope AUC $= 0.736$ vs.\ $0.541$), with Hellenistic sites clustering in coastal lowlands (mean elevation 176\,m vs.\ background 455\,m). However, this signal reflects a broad geographic gradient (coastal lowlands vs.\ mountain interior) rather than fine-grained site-specific signatures. DPL cannot distinguish individual site locations from surrounding coastal terrain, causing positive pseudo-labels to spread across the entire lowland zone. In contrast, Sagalassos's discriminative power comes primarily from historical infrastructure features absent in Cyprus: distance to Hellenistic roads (AUC $= 0.842$) and the Late Antique settlement distance map (AUC $= 0.790$) provide spatially specific signals along road corridors and known settlement zones. The ensemble probability surface (Supplementary Figure~\ref{fig:supp:cyprus_uniform_surfaces}) confirms this: predictions are dominated by coastal proximity gradients, and the mean probability at confirmed site locations (0.183) barely exceeds the landscape mean (0.153). Each fold further trains on only 76 of the 95 site clusters (20\% held out), compounding the already limited positive supervision.

\begin{table}[h]
\centering
\caption{Cyprus Hellenistic uniform site-level k-fold ($r=1$\,m). DL: mean $\pm$ std over 5 folds $\times$ 5 seeds. LAMAP: mean $\pm$ std over 5 folds, 1 seed (deterministic). Bold: best per metric. $^*$\,positive collapse. DPL uses $\tau_{\text{neg,start}}=0.4$, consistency disabled (see text). $^\ddagger$\,LAMAP hyperparameters not tuned for Cyprus; see Supplementary Section~\ref{supp:lamap_params}.}
\label{tab:phase1_cyprus_uniform}
\begin{tabularx}{\textwidth}{l*{2}{>{\centering\arraybackslash}X}}
\toprule
Model & ROC-AUC & Recall \\
\midrule
SL
& $0.494 \pm 0.032$
& $0.001 \pm 0.007$ \\

SL\_Pos
& $0.497 \pm 0.027$
& $^*0.999 \pm 0.007$ \\

DPL
& $0.510 \pm 0.031$
& $0.239 \pm 0.333$ \\

\textbf{LAMAP}$^\ddagger$
& $\mathbf{0.765 \pm 0.041}$
& $\mathbf{0.679 \pm 0.111}$ \\
\bottomrule
\end{tabularx}
\end{table}

\subsubsection{Spatial k-fold (secondary)}

Supplementary Table~\ref{tab:phase1_cyprus_spatial} reports results under spatial k-fold with 4\,km exclusion buffers. Both DL methods show highly asymmetric fold-to-fold variation (DPL: 0.198--0.776; SL: 0.311--0.882), reflecting spatial heterogeneity rather than learned site associations. SL\_Pos retains its collapse.

LAMAP achieves ROC-AUC $= 0.765 \pm 0.041$ (uniform) and $0.678 \pm 0.162$ (spatial), substantially outperforming all deep learning methods. Results may improve with domain-calibrated hyperparameters (Supplementary Section~\ref{supp:lamap_params}).

%% file: sections/07_ablations.tex
We conducted six ablation studies on the Sagalassos Late Antique dataset; full results are in Supplementary Section~\ref{supp:ablation}. Hard-threshold negative pseudolabels are essential: soft-weighted and positive-only variants collapse to Recall\,=\,1.0, while the optimal $\tau_{\text{neg,start}}=0.4$ recovers 97\% of the SL upper-bound ROC-AUC (0.603) and is adopted for Cyprus. Entropy minimisation and the supervised anchor are equally non-negotiable; removing either causes collapse, whereas removing consistency regularisation yields a marginal AUC improvement (0.598 vs.\ 0.592). ResNet-18 is the optimal backbone (AUC 0.612); ResNet-50 underperforms (0.542) due to overfitting under label scarcity, and MobileNetV2 (0.515) is competitive for resource-constrained settings. The LAMAP-derived 295\,m training radius is optimal (AUC 0.600); smaller radii provide insufficient supervision, larger ones introduce noise. DEM features outperform Landsat~9 (AUC 0.598 vs.\ 0.531); DPL prevents collapse on both, unlike SL and SL\_Pos which collapse on Landsat~9. Across positive sampling fractions 0.1--0.5, DPL maintains Recall\,<\,1.0; $\text{pos\_fraction}=0.1$ yields best AUC (0.606).

%% file: sections/08_discussion.tex
\paragraph{The PU framing is empirically necessary.}
SL fails catastrophically under survey-completion evaluation: it misses 85\% of sites on Sagalassos and inverts probability rankings under Cyprus uniform k-fold. Under geographic generalization (Cyprus spatial k-fold), SL's performance varies widely across folds (ROC-AUC range 0.311--0.882), reflecting spatial heterogeneity rather than consistent learning. SL\_Pos collapses consistently across all regimes, predicting positive everywhere. DPL operates between these empirical bounds, recovering 97\% of SL's ROC-AUC while maintaining the discriminative recall that supervised learning cannot achieve in a PU setting.

\paragraph{DPL versus LAMAP: complementary strengths.}
On Sagalassos, DPL achieves better site discovery (+12.5\% F1, +29.3\% recall) while LAMAP leads in probability ranking: PR-AUC 0.985 vs 0.937 and ROC-AUC 0.884 vs 0.601. On Cyprus, DPL nominally operates between the empirical bounds under uniform site-level k-fold (ROC-AUC $= 0.510$) but the margin is not practically meaningful and training collapses in many folds (Section~\ref{sec:phase1}). The feature separability analysis (Supplementary Section~\ref{supp:feature_separability}) identifies the mechanism: DPL's success on Sagalassos depends on historically grounded infrastructure features, specifically distance to Hellenistic roads (AUC $= 0.842$) and a period-specific settlement distance map (AUC $= 0.790$), which provide spatially fine-grained pseudo-label targets. Cyprus lacks equivalent features, leaving only broad terrain gradients (coastal elevation, slope) that create zone-level rather than site-level discrimination. LAMAP's feature-similarity mechanism is robust to this limitation because it measures proximity to known sites directly in feature space, requiring no negative signal whatsoever, and substantially outperforms all deep learning methods (ROC-AUC $= 0.765$ uniform, $0.678$ spatial). This delineates a precondition for DPL: it requires either spatially specific auxiliary features (infrastructure maps, period distance maps) or landscapes where site locations create fine-grained terrain signatures, not merely broad geographic gradients.

\paragraph{Ablation insights: principles for PU learning under label scarcity.}
Three design elements are non-negotiable: entropy minimisation, hard negative thresholding, and a supervised positive anchor; removing any causes collapse. Entropy minimisation is the most critical: without it the model cannot commit to predictions and collapses to the same uniform high-recall behaviour as SL\_Pos. Model capacity must match label scarcity: ResNet-50 overfits while ResNet-18 generalises.

\paragraph{The necessity of negative sampling.}
Pseudo-absence sampling is a known source of bias in predictive modeling. Zhang et al.~\cite{zhang2025interpretable} address this via KDE-based negative sampling, defining pseudo-absence zones from the spatial sphere of influence around known sites. While interpretable, this requires strong prior assumptions about settlement extent that are problematic in heterogeneous landscapes or where no prior settlement model exists. Our approach avoids these priors: the dual-branch network identifies confidently non-site locations from the training signal itself, without encoding spatial distance rules, and generalises to settings where no prior model is available.

\paragraph{Limitations.}
Cyprus lacks field-validated negatives, limiting evaluation to internal cross-validation and forcing each fold to train on only 76 of the 95 site clusters (20\% held out per fold), compounding the already limited positive supervision. The high recall variance (Recall $0.239 \pm 0.333$, Table~\ref{tab:phase1_cyprus_uniform}) illustrates that purely positive-only benchmarks require careful interpretation: without confirmed negatives, training instability cannot be distinguished from spatial heterogeneity. The historical infrastructure features that drive Sagalassos performance also raise a sampling bias concern: sites cluster near Hellenistic roads (AUC $= 0.842$) partly because surveys follow road access, not only because settlement genuinely tracked road networks; DPL may therefore generalise well within surveyed corridors while failing to predict sites in unsurveyed terrain where no road-adjacent bias exists. Confirmed negative labels remain structurally scarce even on Sagalassos (Survey 2: 9 confirmed negatives, 81 positives); randomised field transects are needed to disentangle survey bias from site signal. All predictive surfaces are testable hypotheses, but field validation remains the final arbiter.

%% file: sections/09_conclusion.tex
We frame archaeological predictive modeling as a positive-unlabeled learning task and propose semi-supervised asymmetric dual-branch pseudolabeling, an end-to-end deep learning approach, to learn negatives from data without relying on strong assumptions. In this challenging setting of label scarcity and absent confirmed negatives, we demonstrate that standard supervised learning fails catastrophically when the assumption that unlabeled land is negative is violated, while positive-only supervision collapses to predicting positives everywhere. These baselines establish empirical bounds: DPL operates between them, recovering 97\% of the SL upper-bound ROC-AUC while maintaining a discriminative recall that supervised learning cannot achieve. On Cyprus Hellenistic, a pure PU setting, SL inverts probability rankings, while DPL nominally operates between the empirical bounds.

Across all seven historical periods in the Sagalassos case study, DPL consistently outperforms LAMAP in discovery metrics, while LAMAP retains advantages in probability ranking. In practice, DPL is most effective when period-specific auxiliary features provide spatially fine-grained discrimination; where only coarse terrain gradients are available, as in Cyprus, LAMAP's feature-similarity ranking is the appropriate tool. Together, they offer a richer and more actionable picture of the predictive landscape than either method alone.

This work extends beyond archaeology to the general challenge of learning from incomplete, positively biased labels in spatially structured domains, recurring in ecology, epidemiology, and remote sensing. Asymmetric pseudolabeling offers a tractable path forward without requiring confirmed absences or class-prior estimation. The Cyprus results delineate the boundary of applicability: pseudo-label bootstrapping requires spatially specific discriminative features; where only broad geographic gradients are available, LAMAP's feature-similarity ranking is more appropriate and targeted field survey remains the priority investment.

%% file: sections/supplementary.tex
\section{Implementation Details}
\label{supp:implementation}

Experiments were conducted on an HPC cluster (NVIDIA A100 and P100 GPUs). Models were implemented in PyTorch with TorchGeo~\cite{Stewart_TorchGeo_Deep_Learning_2024}, trained with Adam ($\text{lr}=10^{-4}$, cosine annealing), batch size 32, and focal loss for pseudolabels. Results are reported as mean $\pm$ std over 5 seeds.

\paragraph{DPL hyperparameters.} The pseudolabel temperature is $T=2.0$. The focal loss uses $\alpha=0.02$ and $\gamma=2.0$. Default threshold schedules for Sagalassos: $\tau_{\text{pos,start}}=0.7$, $\tau_{\text{pos,end}}=0.9$, $\tau_{\text{neg,start}}=0.4$, $\tau_{\text{neg,end}}=0.2$, ramped over 20 epochs. The starting threshold $\tau_{\text{neg,start}}=0.4$ is selected via the pseudolabel ablation (Section~\ref{supp:ablation}). For Cyprus Hellenistic, the same $\tau_{\text{neg,start}}=0.4$, $\tau_{\text{neg,end}}=0.2$ configuration is used with consistency regularization disabled.

\section{Additional Results}

\subsection{Metric Trade-offs: Radar Analysis}
\label{supp:radar}
\begin{figure}[h]
\centering
\includegraphics[width=0.7\textwidth]{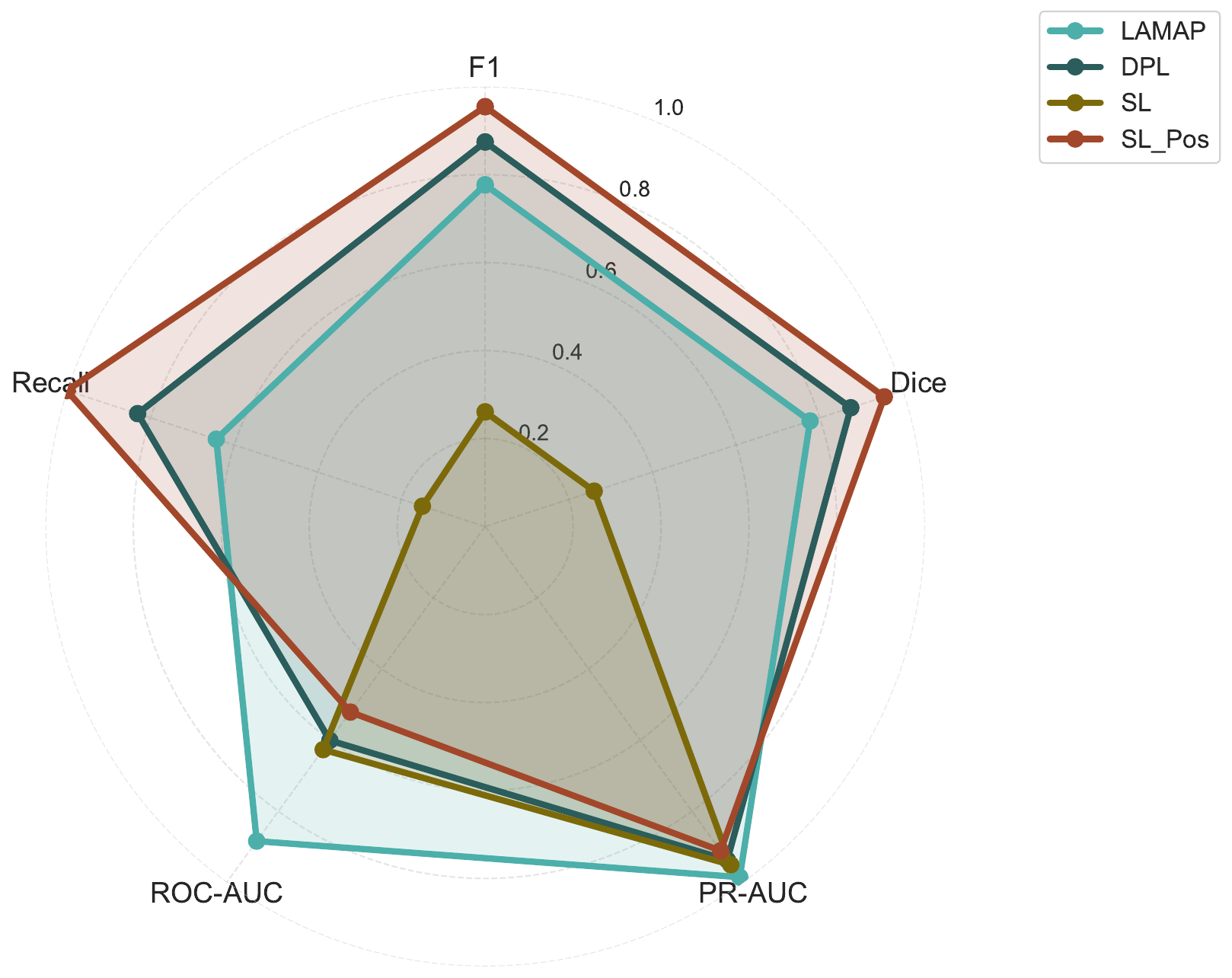}
\caption{Late Antique model comparison across five metrics (F1, Dice, PR-AUC, ROC-AUC, Recall). DPL excels in discovery metrics (F1, Dice, Recall), while LAMAP excels in ranking metrics (PR-AUC, ROC-AUC). SL\_Pos collapses to predicting positive everywhere (Recall=1.0), and SL fails catastrophically.}
\label{fig:supp:phase2_radar}
\end{figure}

Figure \ref{fig:supp:phase2_radar} shows the five-metric trade-off for Late Antique, visualizing the complementary strengths of DPL and LAMAP. DPL shows stronger performance in F1 (0.874), Dice (0.874), and Recall (0.831), the discovery metrics. LAMAP shows stronger performance in PR-AUC (0.985) and ROC-AUC (0.884), the ranking metrics. SL\_Pos shows high F1/Dice from perfect recall (Recall=1.0) but contributes no discriminative value. SL shows near-zero performance across all metrics.

The radar plot volume (normalized area under the polygon) quantifies overall performance: SL\_Pos (0.868), DPL (0.824), LAMAP (0.813), SL (0.450). While SL\_Pos has the largest volume due to inflated F1/Dice from perfect recall (Recall=1.0), this masks its fundamental failure mode.

\subsection{Multi-Period Performance Trends}
\label{supp:period_performances}
Figure \ref{fig:supp:phase2_lineplots} shows F1, PR-AUC, ROC-AUC, and Recall across all seven historical periods in chronological order. Key patterns emerge:

\begin{figure}[h]
\centering
\includegraphics[width=0.45\textwidth]{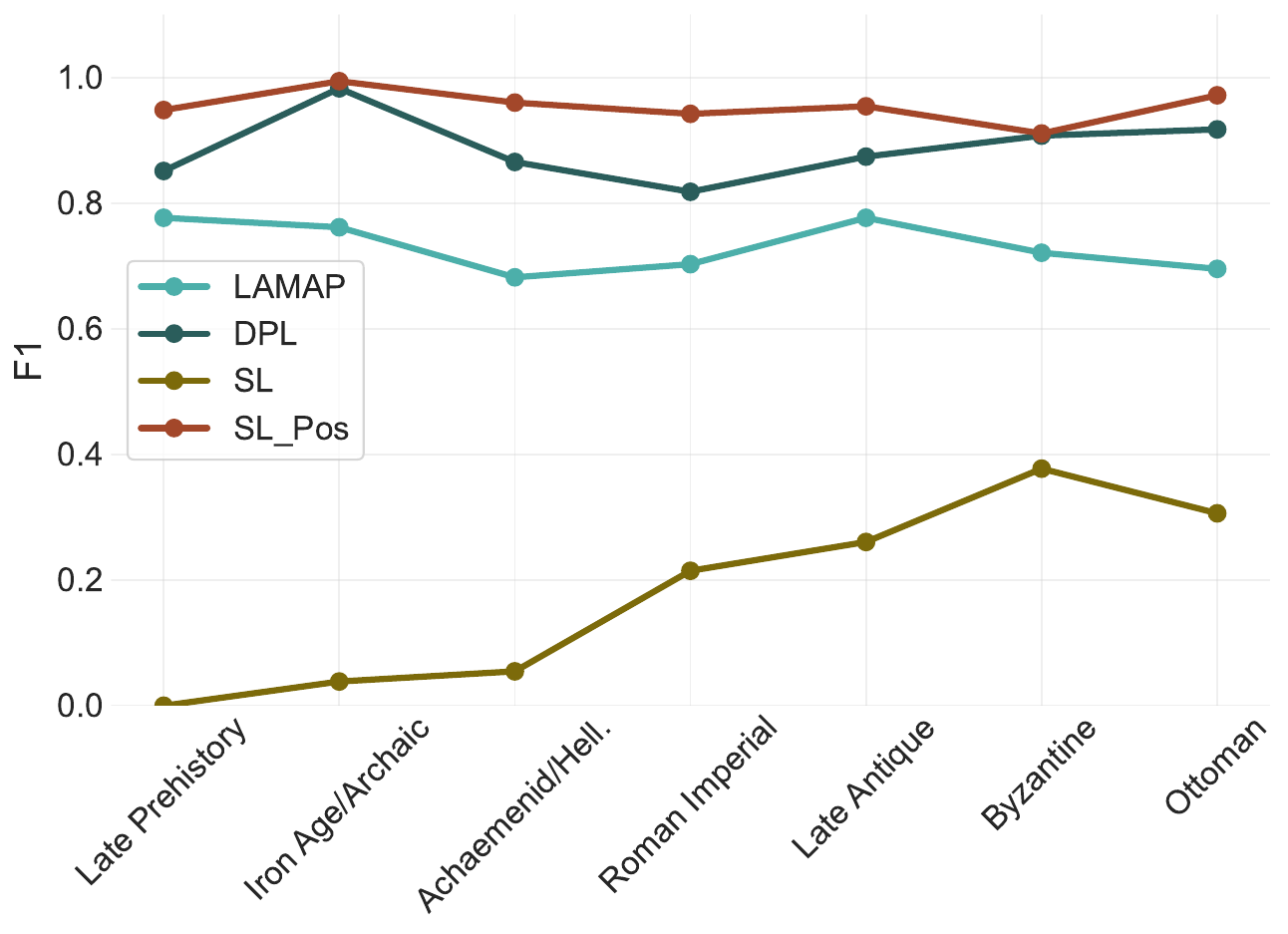}
\includegraphics[width=0.45\textwidth]{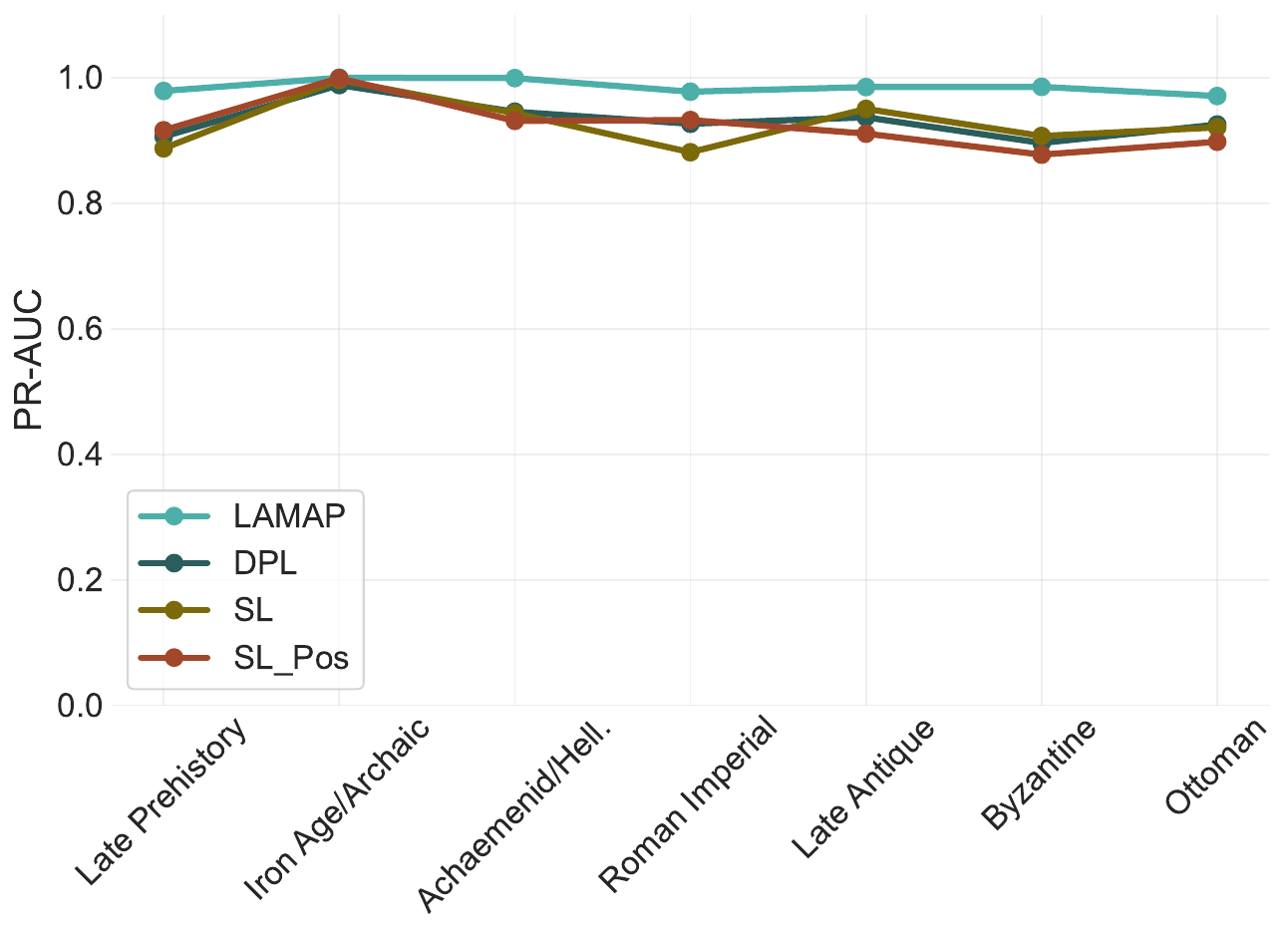}\\
\includegraphics[width=0.45\textwidth]{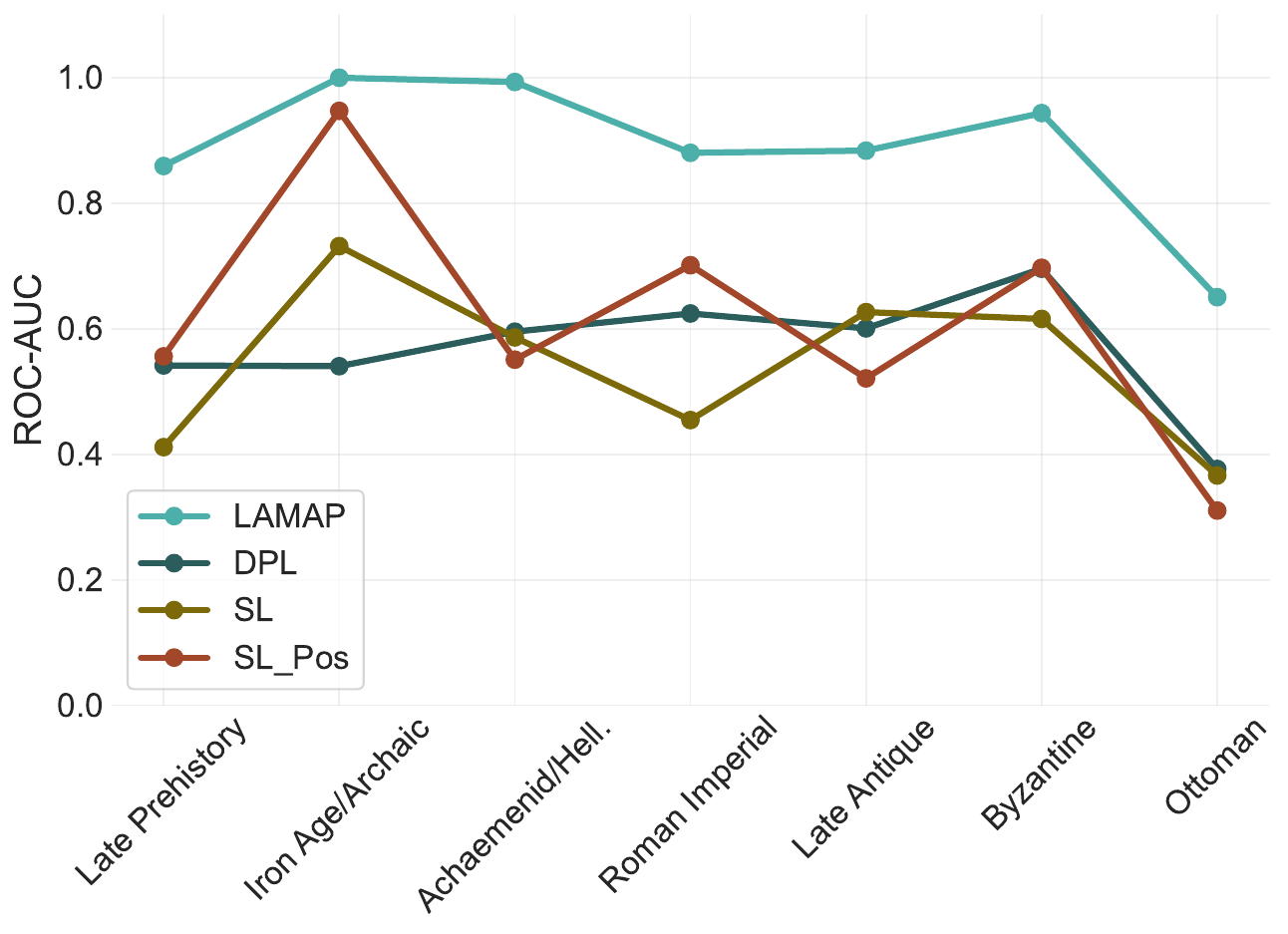}
\includegraphics[width=0.45\textwidth]{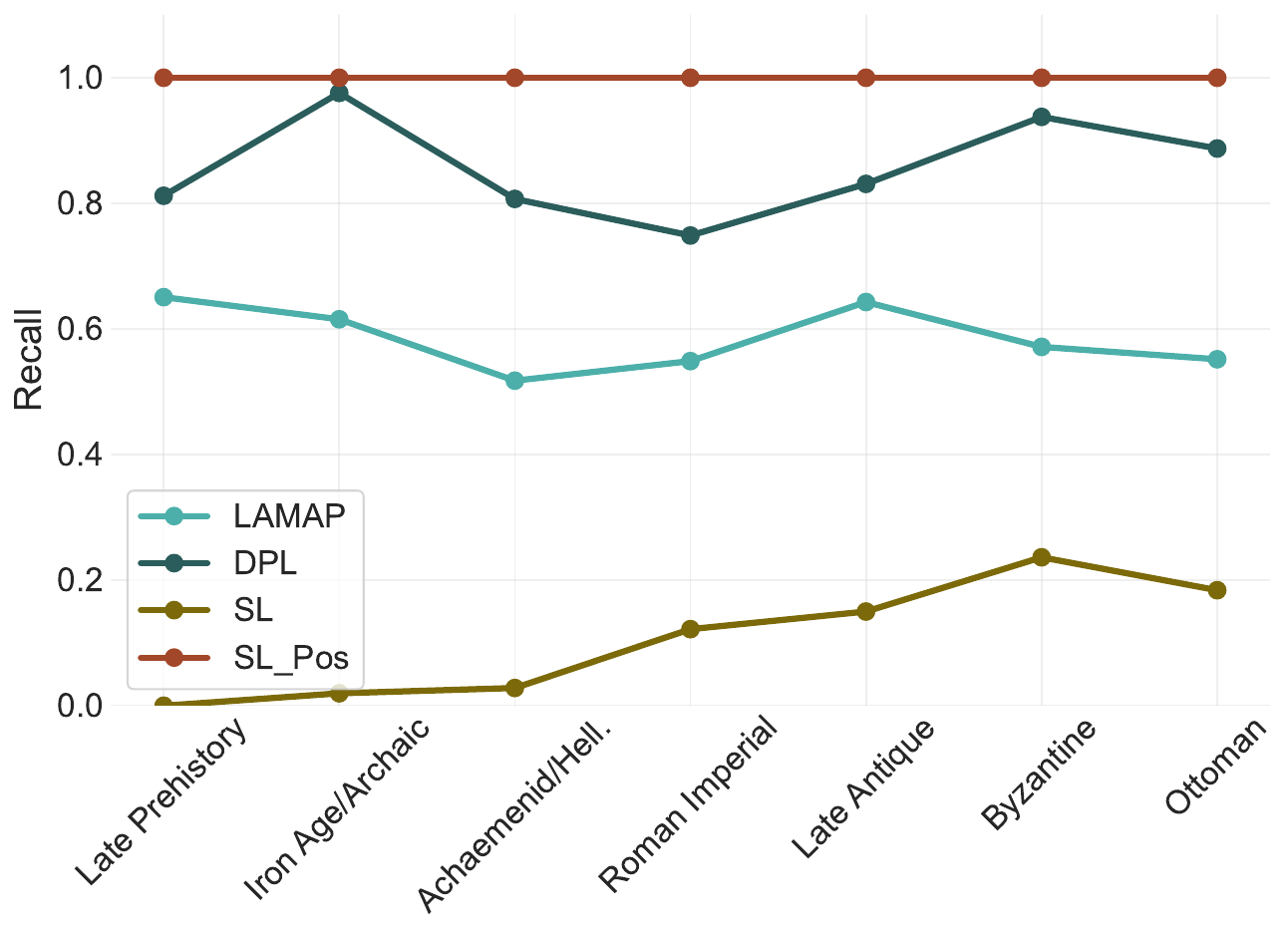}
\caption{Multi-period performance trends across seven historical periods (chronological order): (a) F1 Score, (b) PR-AUC, (c) ROC-AUC, (d) Recall. DPL consistently outperforms LAMAP in F1 and Recall across all periods. LAMAP maintains advantages in PR-AUC and ROC-AUC. SL\_Pos collapses to Recall=1.0 (model collapse), and SL shows near-zero performance (catastrophic failure).}
\label{fig:supp:phase2_lineplots}
\end{figure}

\paragraph{Discovery metrics (F1, Recall).} DPL consistently above LAMAP across all periods, with the F1 gap widest in Ottoman (32\% advantage) and narrowest in Late Prehistory (10\%).

\paragraph{Ranking metrics (PR-AUC, ROC-AUC).} LAMAP consistently above DPL across all periods, reflecting its explicit probability modeling advantage.

\paragraph{SL\_Pos.} Flat line at Recall=1.0 across all periods (model collapse).

\paragraph{SL.} Near-zero performance across all periods (catastrophic failure).

The consistency of these patterns across diverse historical contexts---with different site densities, landscape configurations, and cultural periods---demonstrates that our findings are not artifacts of a specific time period but reflect fundamental methodological differences.

\subsection{Surface predictions across all archaeological periods}

\begin{figure}[htbp]
  \centering
  \includegraphics[width=\textwidth]{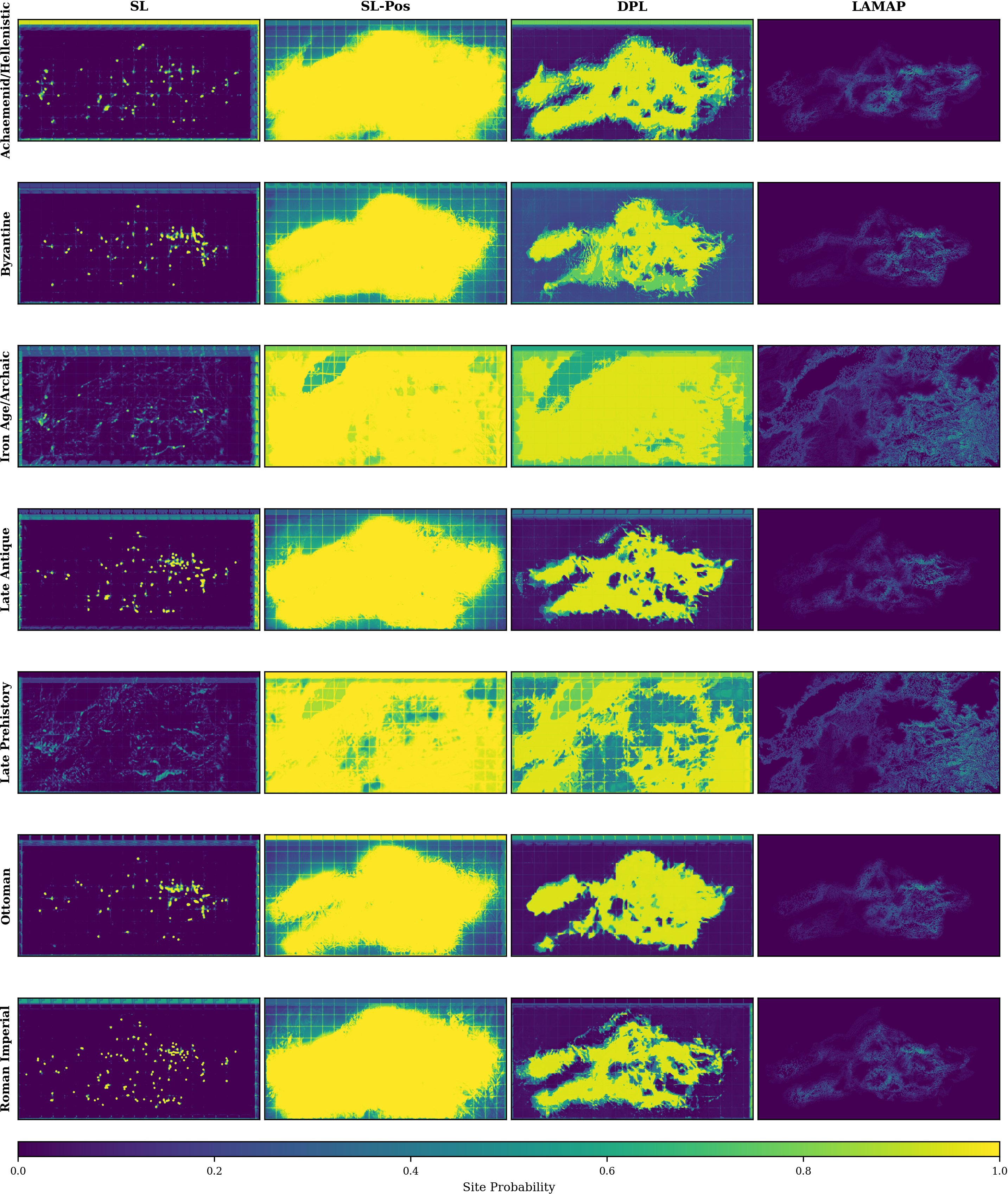}
  \caption{Surface probability predictions for different models and historical periods. Models: SL, SL\_Pos, DPL, LAMAP. Periods: Achaemenid/Hellenistic, Byzantine, Iron Age--Archaic, Late Antique, Late Prehistory, Ottoman, Roman Imperial. Probability maps show mean prediction across 5 random seeds.}
  \label{fig:supp:surface_grid}
\end{figure}

Figure~\ref{fig:supp:surface_grid} visualizes ensemble probability surfaces across all seven historical periods, enabling qualitative comparison of model behaviors.

\subsection{DPL surface prediction variance for Late Antique}

\begin{figure}[htbp]
  \centering
  \includegraphics[width=\textwidth]{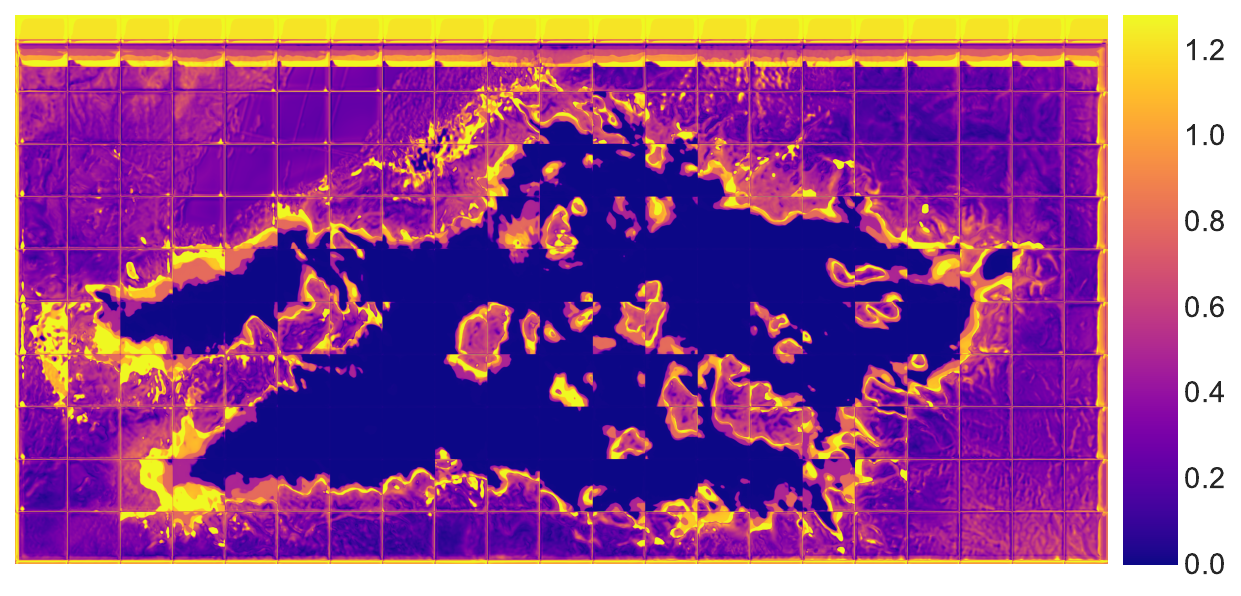}
  \caption{Surface probability variance of DPL for Late Antique. The map shows variance prediction across 5 random seeds.}
  \label{fig:supp:variance}
\end{figure}

Figure~\ref{fig:supp:variance} quantifies prediction uncertainty across five random seeds, highlighting regions where the model is most sensitive to initialization.

\section{Ablations}
\label{supp:ablation}
We explore the pseudolabel strategy, DPL loss components, training label radius, backbone architecture, end-to-end feature learning, and positive sampling fraction. If not otherwise specified, we use a ResNet-18 backbone, a training radius of 295\,m, and a positive sampling fraction of 0.1. All experiments train and evaluate on the Sagalassos Late Antique dataset, which provides the most label-rich PNU evaluation setting.

\subsection{Pseudolabel Strategy}
This ablation investigates the added benefit of negative pseudolabels and the optimal strategy for their generation.
We compare supervised SL and SL\_Pos against multiple DPL variants: varying the initial negative pseudolabel threshold $\tau_{\text{neg,start}} \in \{0.2, 0.4, 0.6\}$ (with $\tau_{\text{neg,end}}=0.2$ fixed), soft-weighted negatives DPL (soft neg), and DPL (pos only), which relies solely on positive pseudolabels. Table \ref{tab:neg_ablation} presents the results. Our default setting of DPL ($\tau_{\text{neg,start}}=0.4$) is optimal, achieving the best ROC-AUC (0.603). Relying only on positive pseudolabels, or using soft negative pseudolabels, causes the model to collapse. All hard-threshold variants successfully learn negatives. However, the start threshold matters: too permissive ($\tau_{\text{neg,start}}=0.2$, no annealing) and ROC-AUC drops (0.571); too conservative ($\tau_{\text{neg,start}}=0.6$) and it also drops (0.559). The collapse of the soft-weighted variant indicates that confidence-based weighting alone is insufficient; a hard, threshold-based negative signal is required.

\begin{table}[h]
\centering
\caption{Effect of initial negative pseudolabel threshold $\tau_{\text{neg,start}}$ on Sagalassos Late Antique ($\tau_{\text{neg,end}}=0.2$ fixed throughout). Results reported as mean $\pm$ std. Bold indicates best performance per metric. $^*$ indicates positive collapse. $^{\dagger}$Results use a different seed set from main Table~\ref{tab:phase2_late_antique_r1}.}
\label{tab:neg_ablation}
\begin{tabularx}{\textwidth}{l*{4}{>{\centering\arraybackslash}X}}
\toprule
Config & ROC-AUC & F1 & Recall & Precision \\
\midrule
SL (upper)
& $\mathbf{0.619 \pm 0.093}$
& $0.257 \pm 0.016$
& $0.148 \pm 0.011$
& $\mathbf{1.000 \pm 0.000}$ \\

DPL ($\tau_{\text{neg,start}}=0.2$)
& $0.571 \pm 0.053$
& $0.879 \pm 0.058$
& $0.838 \pm 0.108$
& $0.932 \pm 0.008$ \\

$\mathbf{DPL\ (\tau_{\text{neg,start}}=0.4)}$
& $0.603 \pm 0.052$
& $\mathbf{0.909 \pm 0.039}$
& $\mathbf{0.890 \pm 0.072}$
& $0.930 \pm 0.004$ \\

DPL ($\tau_{\text{neg,start}}=0.6$)
& $0.559 \pm 0.096$
& $0.886 \pm 0.046$
& $0.852 \pm 0.082$
& $0.925 \pm 0.007$ \\

DPL (soft neg)
& $0.538 \pm 0.162$
& $0.955 \pm 0.000$
& $^*1.000 \pm 0.000$
& $0.913 \pm 0.000$ \\

DPL (pos only)
& $0.413 \pm 0.095$
& $0.955 \pm 0.000$
& $^*1.000 \pm 0.000$
& $0.913 \pm 0.000$ \\

SL\_Pos (lower)
& $0.528 \pm 0.021$
& $0.955 \pm 0.000$
& $^*1.000 \pm 0.000$
& $0.913 \pm 0.000$ \\
\bottomrule
\end{tabularx}
\end{table}

\subsection{Loss Components}
We assess the contribution of each DPL loss component (Equation \ref{eq:dpl_full_loss}) by comparing the full model against variants with individual terms removed, alongside the supervised bounds SL and SL\_Pos. Results are shown in Table \ref{tab:loss_components}. Entropy regularization is critical: removing it causes model collapse (Recall = 1.0, same as SL\_Pos), indicating the model can no longer distinguish negatives from unlabeled regions. Relying solely on pseudolabels without any supervised anchor also collapses, confirming the supervised term stabilises training. The consistency term has minimal effect; removing it achieves slightly higher ROC-AUC (0.598) than full DPL (0.592), suggesting it provides limited benefit in this PU setting. DPL matches or exceeds SL's ROC-AUC (0.592 vs.\ 0.578$^{\dagger}$) while maintaining discriminative F1 (0.866 vs.\ 0.262 for SL).

\begin{table}[h]
\centering
\caption{Loss component impact comparison. Results reported as mean $\pm$ std. Bold indicates best performance per metric. $^*$ indicates positive collapse. $^{\dagger}$Results use a different seed set from main Table~\ref{tab:phase2_late_antique_r1}.}
\label{tab:loss_components}
\begin{tabularx}{\textwidth}{l*{4}{>{\centering\arraybackslash}X}}
\toprule
Config & ROC-AUC & F1 & Recall & Precision \\
\midrule
SL (upper)
& $0.578 \pm 0.048$
& $0.262 \pm 0.015$
& $0.152 \pm 0.010$
& $\mathbf{0.967 \pm 0.013}$ \\

$\mathbf{DPL\ (full)}$
& $0.592 \pm 0.061$
& $0.866 \pm 0.061$
& $0.815 \pm 0.107$
& $0.930 \pm 0.005$ \\

DPL (-consistency)
& $\mathbf{0.598 \pm 0.036}$
& $\mathbf{0.895 \pm 0.052}$
& $\mathbf{0.868 \pm 0.096}$
& $0.929 \pm 0.003$ \\

DPL (-entropy)
& $0.520 \pm 0.075$
& $0.955 \pm 0.000$
& $^*1.000 \pm 0.000$
& $0.913 \pm 0.000$ \\

DPL (pseudo only)
& $0.444 \pm 0.117$
& $0.957 \pm 0.005$
& $^*1.000 \pm 0.000$
& $0.917 \pm 0.009$ \\

SL\_Pos (lower)
& $0.528 \pm 0.029$
& $0.955 \pm 0.000$
& $^*1.000 \pm 0.000$
& $0.913 \pm 0.000$ \\
\bottomrule
\end{tabularx}
\end{table}

\subsection{End-to-End Training}
DEM-derived features undergo a preprocessing pipeline (slope, curvature, hydrological proximity). This ablation compares DEM features with raw Landsat~9 multispectral imagery, for which we require no preprocessing beyond feature normalisation and padding. This satellite imagery provides nine multispectral bands capturing surface reflectance properties \cite{MASEK2020111968}, enabling end-to-end learning without handcrafted features. Results are shown in Table~\ref{tab:landsat_ablation}. DPL trained on DEM achieves a higher ROC-AUC (0.598) compared to DPL trained on Landsat~9 (0.531), indicating that topographic features are more informative than spectral reflectance for detecting archaeological sites in this region. However, DPL prevents collapse on both modalities: DPL + Landsat~9 maintains high F1 (0.891, Recall = 0.874), while SL trained on Landsat~9 misses most sites (Recall = 0.150, F1 = 0.261). This demonstrates that DPL's pseudolabel mechanism is effective regardless of feature type.

\begin{table}[h]
\centering
\caption{Landsat~9 vs.\ DEM feature comparison on Sagalassos Late Antique ($r=1$\,m, 5 seeds, mean $\pm$ std). DPL prevents model collapse regardless of feature type, while SL collapses on Landsat features.}
\label{tab:landsat_ablation}
\begin{tabularx}{\textwidth}{l*{4}{>{\centering\arraybackslash}X}}
\toprule
Config & ROC-AUC & F1 & Recall & Precision \\
\midrule
\textbf{DPL + DEM}
& $\mathbf{0.598 \pm 0.043}$
& $0.882 \pm 0.050$
& $0.843 \pm 0.092$
& $\mathbf{0.929 \pm 0.002}$ \\

DPL + Landsat~9
& $0.531 \pm 0.059$
& $\mathbf{0.891 \pm 0.061}$
& $\mathbf{0.874 \pm 0.105}$
& $0.915 \pm 0.011$ \\

SL + Landsat~9
& $0.535 \pm 0.035$
& $0.261 \pm 0.021$
& $0.150 \pm 0.014$
& $1.000 \pm 0.000$ \\

SL + Landsat~9 (no hist)
& $0.531 \pm 0.036$
& $0.257 \pm 0.016$
& $0.148 \pm 0.011$
& $1.000 \pm 0.000$ \\
\bottomrule
\end{tabularx}
\end{table}

\subsection{Label Radius}
We investigated the effect of the training label radius, following the original LAMAP procedure, which inflated each site label by a radius of 295\,m. Table \ref{tab:radius_effect_r1} shows results for training radii of 1, 295, and 500\,m, all evaluated at $r=1$\,m. The optimal radius is 295\,m (ROC-AUC = 0.600). The $r=1$\,m model (0.542 AUC) sees too few positive pixels per patch, leading to weak supervision. Training with $r=500$\,m achieves the lowest AUC (0.526), as the label becomes too noisy, hurting the model's ability to learn precise site features. Our default $r=295$\,m therefore provides sufficient positive pixels for stable training while avoiding excess label noise.

\begin{table}[h]
\centering
\caption{Training radius effect evaluated at $r=1$\,m. Results are mean $\pm$ std. Bold indicates best performance per metric.}
\label{tab:radius_effect_r1}
\begin{tabularx}{\textwidth}{l*{4}{>{\centering\arraybackslash}X}}
\toprule
Training radius & ROC-AUC & F1 & Recall & Precision \\
\midrule
$r = 1$m
& $0.542 \pm 0.085$
& $0.867 \pm 0.089$
& $0.826 \pm 0.156$
& $0.926 \pm 0.007$ \\

$\mathbf{r = 295}$m
& $\mathbf{0.600 \pm 0.054}$
& $0.887 \pm 0.068$
& $0.855 \pm 0.116$
& $\mathbf{0.930 \pm 0.005}$ \\

$r = 500$m
& $0.526 \pm 0.060$
& $\mathbf{0.908 \pm 0.038}$
& $\mathbf{0.893 \pm 0.069}$
& $0.926 \pm 0.005$ \\
\bottomrule
\end{tabularx}
\end{table}

\subsection{Architecture}
We investigated how backbone architecture affects learning, comparing ResNet-18 against a larger backbone (ResNet-50) and a lightweight alternative (MobileNetV2). Our results, presented in Table \ref{tab:backbone_comparison}, show that the larger model underperforms ResNet-18 (ROC-AUC = 0.542), suggesting that additional capacity leads to overfitting on sparse labels. MobileNetV2 achieves similar F1 (0.863) but lower AUC (0.515), and may serve as an alternative for resource-constrained deployment or fast prototyping.

\begin{table}[h]
\centering
\caption{Backbone comparison. Results reported as mean $\pm$ std. Bold indicates best performance per metric.}
\label{tab:backbone_comparison}
\begin{tabularx}{\textwidth}{l*{4}{>{\centering\arraybackslash}X}}
\toprule
Backbone & ROC-AUC & F1 & Recall & Precision \\
\midrule
ResNet-18
& $\mathbf{0.612 \pm 0.039}$
& $\mathbf{0.883 \pm 0.065}$
& $\mathbf{0.848 \pm 0.110}$
& $\mathbf{0.930 \pm 0.005}$ \\

ResNet-50
& $0.542 \pm 0.058$
& $0.809 \pm 0.079$
& $0.726 \pm 0.120$
& $0.924 \pm 0.010$ \\

MobileNetV2
& $0.515 \pm 0.087$
& $0.863 \pm 0.048$
& $0.812 \pm 0.081$
& $0.924 \pm 0.009$ \\
\bottomrule
\end{tabularx}
\end{table}

\subsection{Positive Sampling Fraction}
We investigated the influence of the positive sampling fraction (default = 0.1), i.e., to what extent the positive-to-background patch-sampling ratio affects training. Results are shown in Table \ref{tab:pos_fraction}. All DPL variants maintain Recall $<$ 1.0 (range 0.848--0.888), confirming that DPL does not collapse at any tested fraction. DPL's ROC-AUC degrades above a sampling fraction of 0.1: the drop from 0.606 (frac = 0.1) to 0.556 (frac = 0.3) and 0.555 (frac = 0.5) suggests that higher positive fractions over-weight positive supervision and reduce the discriminative signal from pseudolabels. SL\_Pos collapses regardless of sampling fraction. SL's precision--recall trade-off shifts strongly with positive fraction: higher values force SL to predict more positives, increasing Recall (0.145 $\to$ 0.388) but degrading precision (1.000 $\to$ 0.915). This reflects sampling imbalance directly, not a model improvement; AUC stays flat or declines.

\begin{table}[h]
\centering
\small
\setlength{\tabcolsep}{4pt}
\caption{Effect of positive sampling fraction. Results reported as mean $\pm$ std. Bold indicates best performance per metric. $^*$ indicates positive collapse.}
\label{tab:pos_fraction}
\resizebox{\textwidth}{!}{%
\begin{tabular}{l c *{4}{c}}
\toprule
Model & pos\_fraction & ROC-AUC & F1 & Recall & Precision \\
\midrule

\multirow{4}{*}{SL}
& 0.1 & 0.590 $\pm$ 0.099 & 0.254 $\pm$ 0.015 & 0.145 $\pm$ 0.010 & 1.000 $\pm$ 0.000 \\
& 0.2 & 0.592 $\pm$ 0.048 & 0.317 $\pm$ 0.032 & 0.190 $\pm$ 0.022 & 0.952 $\pm$ 0.028 \\
& 0.3 & 0.545 $\pm$ 0.056 & 0.495 $\pm$ 0.087 & 0.345 $\pm$ 0.084 & 0.909 $\pm$ 0.029 \\
& 0.5 & 0.527 $\pm$ 0.017 & 0.544 $\pm$ 0.039 & 0.388 $\pm$ 0.036 & 0.915 $\pm$ 0.022 \\

\addlinespace

\multirow{4}{*}{SL\_Pos}
& 0.1 & 0.523 $\pm$ 0.025 & 0.955 $\pm$ 0.000 & $^*$1.000 $\pm$ 0.000 & 0.913 $\pm$ 0.000 \\
& 0.2 & 0.501 $\pm$ 0.038 & 0.955 $\pm$ 0.000 & $^*$1.000 $\pm$ 0.000 & 0.913 $\pm$ 0.000 \\
& 0.3 & 0.512 $\pm$ 0.017 & 0.955 $\pm$ 0.000 & $^*$1.000 $\pm$ 0.000 & 0.913 $\pm$ 0.000 \\
& 0.5 & 0.529 $\pm$ 0.027 & 0.955 $\pm$ 0.000 & $^*$1.000 $\pm$ 0.000 & 0.913 $\pm$ 0.000 \\

\addlinespace

\multirow{4}{*}{DPL}
& \textbf{0.1} & \textbf{0.606 $\pm$ 0.052} & 0.885 $\pm$ 0.044 & 0.848 $\pm$ 0.080 & 0.929 $\pm$ 0.005 \\
& 0.2 & 0.599 $\pm$ 0.103 & \textbf{0.907 $\pm$ 0.047} & \textbf{0.888 $\pm$ 0.086} & \textbf{0.930 $\pm$ 0.003} \\
& 0.3 & 0.556 $\pm$ 0.114 & 0.890 $\pm$ 0.038 & 0.860 $\pm$ 0.064 & 0.925 $\pm$ 0.009 \\
& 0.5 & 0.555 $\pm$ 0.071 & 0.890 $\pm$ 0.048 & 0.860 $\pm$ 0.086 & 0.925 $\pm$ 0.007 \\

\bottomrule
\end{tabular}
}
\end{table}

\section{Cyprus Hellenistic}

\subsection{K-Fold Validation Strategies}
\label{supp:kfold}

We evaluate Cyprus Hellenistic under two complementary k-fold strategies (Section~\ref{sec:experimental_design}): uniform site-level k-fold (primary) and spatial k-fold with 4\,km buffer zones (secondary).

\subsubsection{Uniform Site-Level K-Fold}
\label{supp:kfold_uniform}

\begin{figure}[h]
\centering
\includegraphics[width=0.48\textwidth]{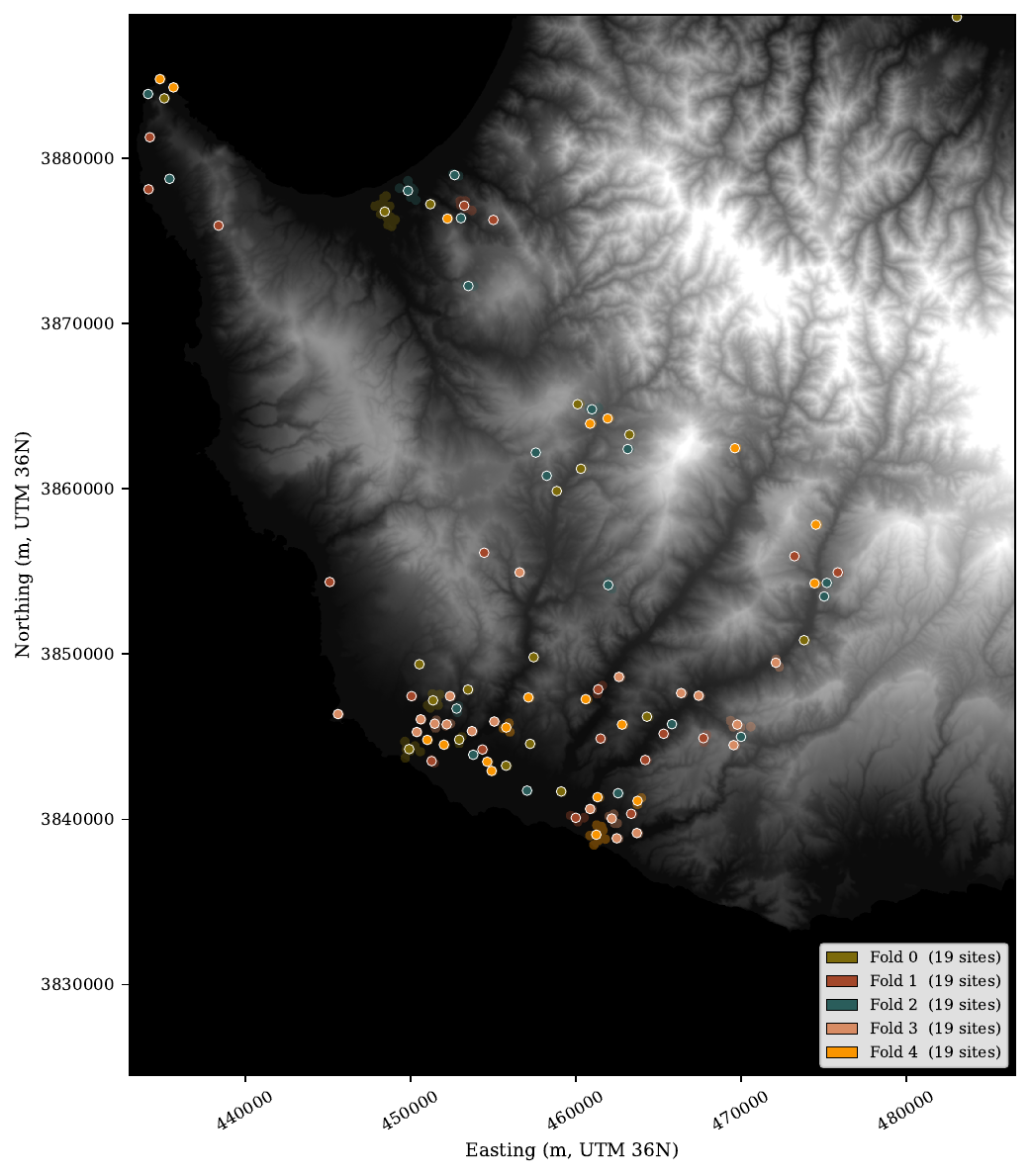}
\includegraphics[width=0.48\textwidth]{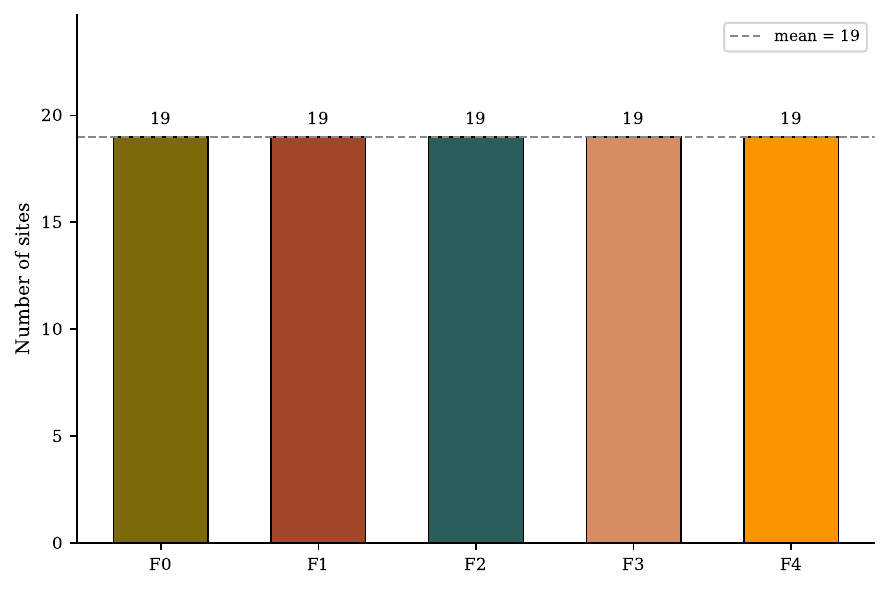}
\caption{Uniform site-level k-fold for Cyprus Hellenistic. (a)~Geographic map with 95 site clusters colour-coded by fold assignment. (b)~Cluster count per fold (balanced at 19 sites each).}
\label{fig:supp:kfold_uniform}
\end{figure}

Figure~\ref{fig:supp:kfold_uniform} illustrates the uniform k-fold strategy. 95 Hellenistic site clusters (formed from 167 sites at the 295\,m training radius) are randomly assigned to 5 balanced folds (19 sites per fold). All sites within a connected-component cluster remain in the same fold, preventing data leakage. Panel (a) shows the geographic distribution; panel (b) confirms balanced fold sizes by design.

\subsubsection{Spatial K-Fold with 4\,km Buffers}
\label{supp:kfold_spatial}

\begin{figure}[h]
\centering
\begin{minipage}{0.48\textwidth}
\centering
\includegraphics[width=\textwidth]{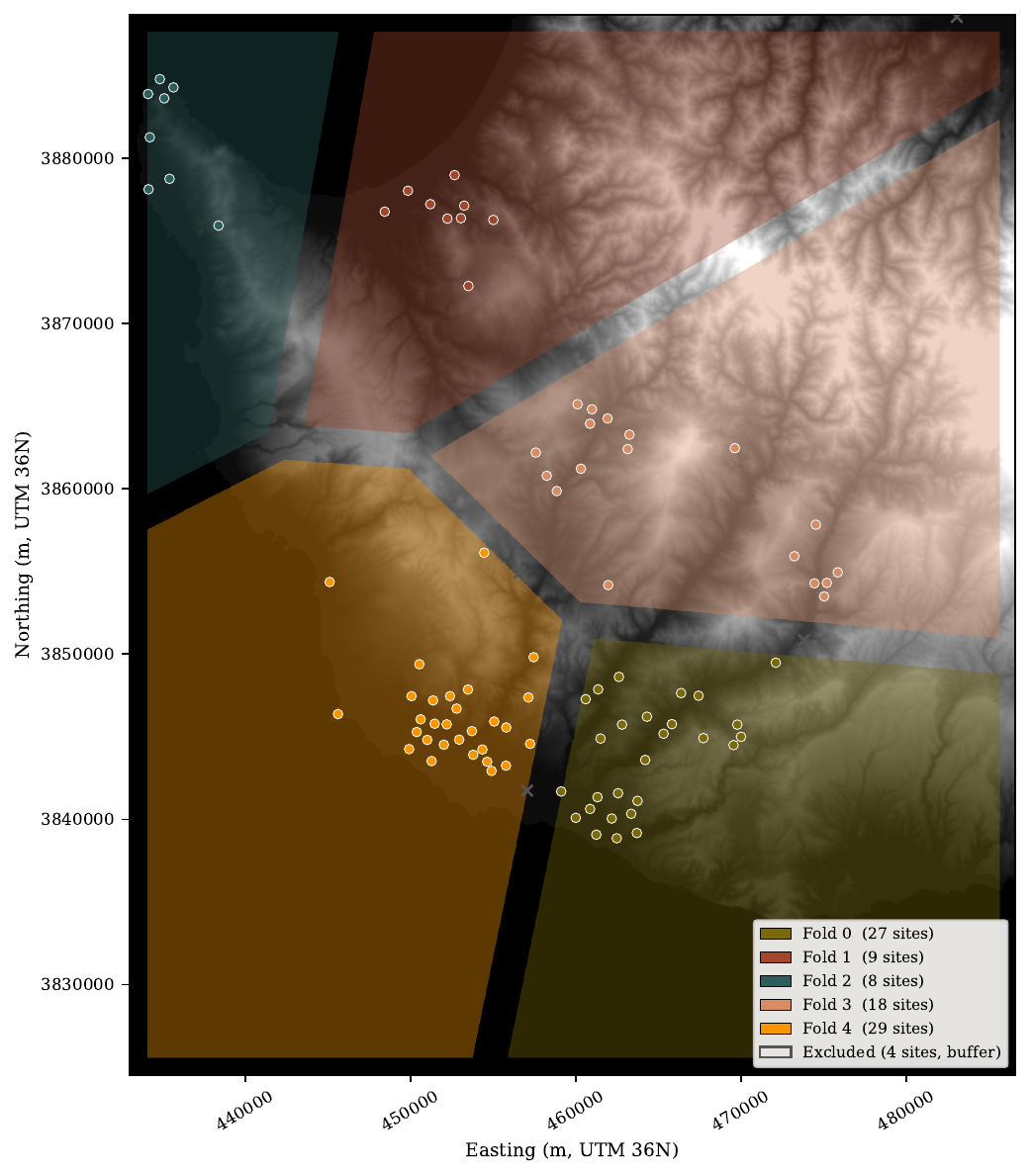}
\end{minipage}
\hfill
\begin{minipage}{0.48\textwidth}
\centering
\includegraphics[width=\textwidth]{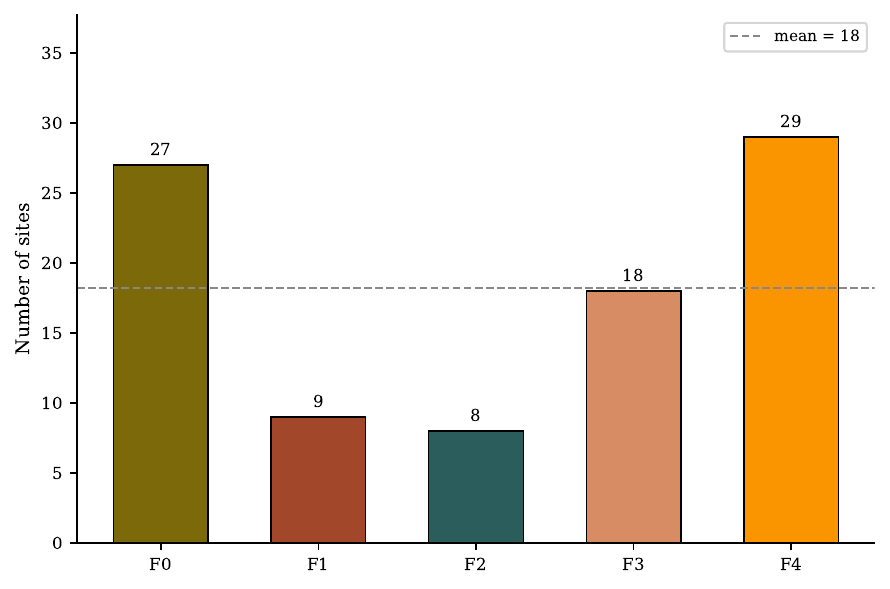} \\[0.3em]
\includegraphics[width=\textwidth]{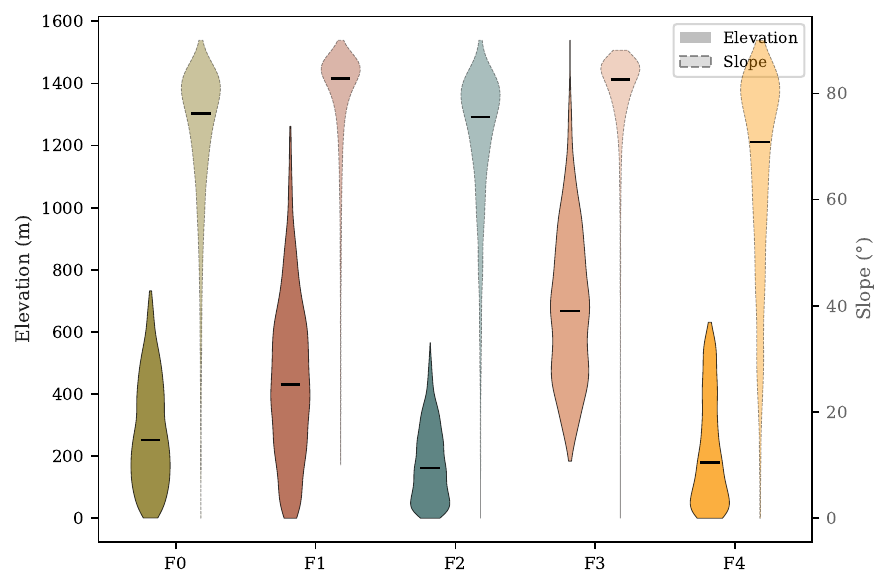}
\end{minipage}
\caption{Spatial k-fold with 4\,km buffer zones for Cyprus Hellenistic. (a)~Geographic map: fold regions colour-coded on the DEM background; coloured dots = cluster centroids; grey crosses = buffer-excluded clusters. (b)~Cluster count per fold (mean = 19, range 8--31). (c)~Elevation (solid violins) and slope (hatched violins) distributions per fold; similar distributions confirm limited covariate shift.}
\label{fig:supp:kfold_spatial}
\end{figure}

Figure~\ref{fig:supp:kfold_spatial} illustrates the spatial k-fold strategy. Fold boundaries are determined by K-means++ clustering of connected-component centroids (95 clusters from 167 Hellenistic sites at the 295\,m training radius). A 4\,km buffer zone around fold boundaries excludes clusters near edges (4 sites excluded, 4.2\%), creating geographic gaps that test whether models can generalize to held-out spatial regions. Panel (a) shows the spatial partition; panel (b) shows fold sizes are moderately imbalanced (range 8--31 clusters), reflecting uneven site density. Panel (c) confirms elevation and slope distributions are broadly similar across folds, indicating limited covariate shift.

\subsection{Probability Surfaces for Uniform K-Fold}

\begin{figure}[h]
\centering
\includegraphics[width=\textwidth]{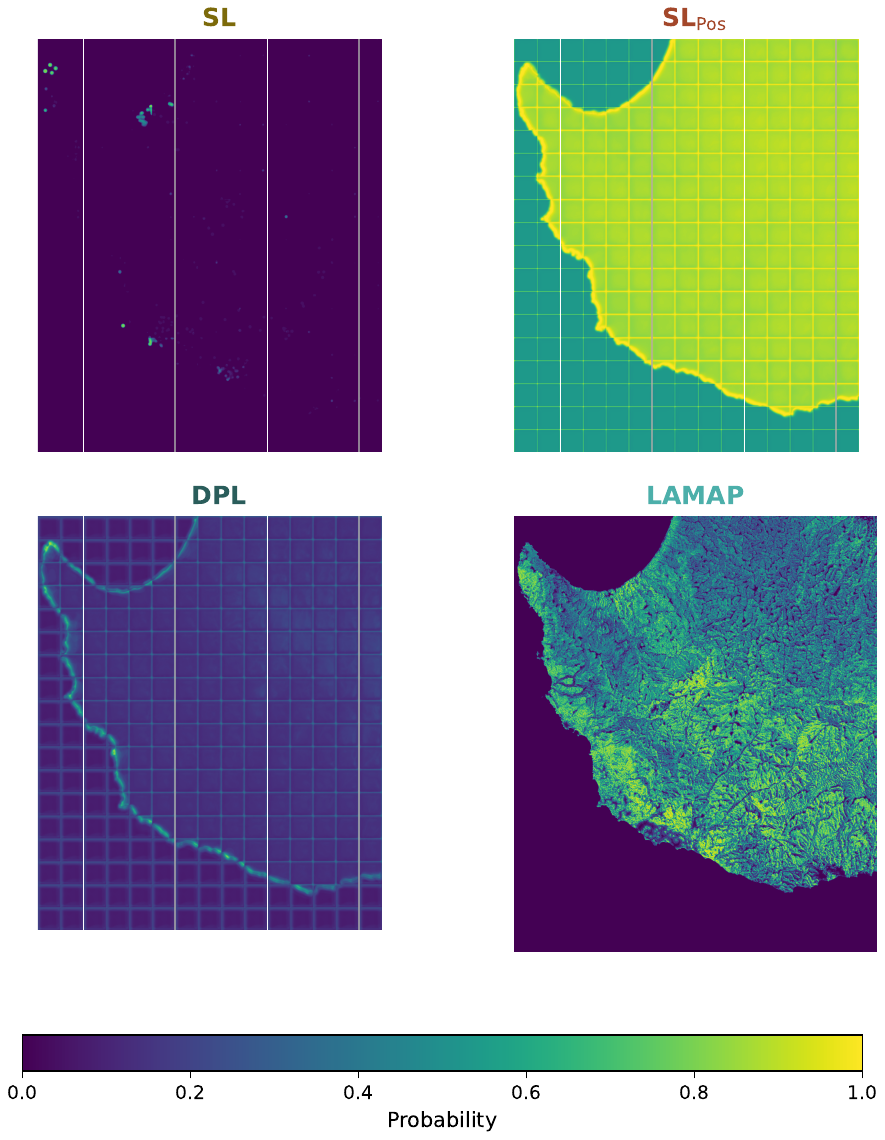}
\caption{Cyprus Hellenistic uniform site-level k-fold ensemble probability surfaces (5 seeds $\times$ 5 folds, 25 predictions averaged per model). All panels share a common viridis colour scale (0--1, purple to yellow). (a)~SL: uniformly low probability, closed-world collapse. (b)~SL\_Pos: uniformly high probability, positive-only collapse. (c)~DPL: intermediate predictions with coastal gradients and high variance. (d)~LAMAP: feature-similarity based, robust to PU uncertainty.}
\label{fig:supp:cyprus_uniform_surfaces}
\end{figure}

Figure~\ref{fig:supp:cyprus_uniform_surfaces} shows ensemble probability surfaces for the uniform site-level k-fold evaluation. Each panel averages predictions across all 5 seeds $\times$ 5 folds (25 predictions), with test-fold sites masked out during training for each fold.

\paragraph{SL (panel a)} produces uniformly low probabilities across the landscape (purple), reflecting the closed-world collapse: treating unlabeled background as negatives during training causes SL to score confirmed sites lower than surrounding terrain (ROC-AUC $= 0.494 \approx 0.5$, Recall $= 0.001$).

\paragraph{SL\_Pos (panel b)} produces uniformly high probabilities (yellow-green), confirming positive-only collapse: without negative supervision, the model predicts sites everywhere (Recall $\approx 1.0$, ROC-AUC $= 0.497$).

\paragraph{DPL (panel c)} shows intermediate predictions dominated by coastal terrain gradients rather than site-specific features. The mean probability at confirmed site locations (0.153) barely exceeds the landscape mean (0.153), and high fold-to-fold variance reveals training instability (Recall $= 0.239 \pm 0.333$, with some folds collapsing toward SL\_Pos behavior and others toward near-zero predictions).

\paragraph{LAMAP (panel d)} demonstrates robust performance via feature-space similarity, substantially outperforming all deep learning methods (ROC-AUC $= 0.765$, Recall $= 0.679$). LAMAP avoids pseudo-label instability by relying exclusively on similarity to known training sites, without requiring negative signal.

\subsection{Feature Separability Analysis}
\label{supp:feature_separability}

To understand the divergence in DPL performance between the two datasets, we compute per-feature site-vs-background separability for Sagalassos Late Antique and Cyprus Hellenistic. For each feature band, we compare pixel values at site-labeled locations (label $> 0$, dilated to 295\,m radius) against background pixels (label $= 0$), reporting Mann-Whitney AUC (equivalent to the probability that a randomly drawn site pixel scores higher than a randomly drawn background pixel; 0.5 = chance) and Cohen's $d$ (pooled-standard-deviation normalised mean difference).

\begin{table}[h]
\centering
\caption{Feature separability: site pixels vs.\ background pixels (Mann-Whitney AUC and Cohen's $d$). Sagalassos Late Antique uses 8 input channels (5 DEM-derived + 3 historical infrastructure); Cyprus Hellenistic uses 5 DEM-derived channels only. AUC $= 0.5$ indicates no discriminative power; values are flipped to always report $\geq 0.5$.}
\label{tab:feature_separability}
\begin{tabularx}{\textwidth}{l l *{2}{>{\centering\arraybackslash}X}}
\toprule
Dataset & Feature & AUC & Cohen's $d$ \\
\midrule
\multirow{8}{*}{Sagalassos Late Antique}
& DEM                        & 0.540 & \phantom{$-$}0.09 \\
& Aspect                     & 0.501 & $-$0.05 \\
& Slope                      & 0.541 & \phantom{$-$}0.05 \\
& Curvature                  & 0.529 & \phantom{$-$}0.09 \\
& Water dist.                & 0.718 & \phantom{$-$}0.80 \\
& Dist.\ Hellen.\ road       & \textbf{0.842} & $-$1.33 \\
& Dist.\ Roman road          & 0.547 & $-$0.24 \\
& Late Antique dist.\ map    & \textbf{0.790} & $-$1.13 \\
\midrule
\multirow{5}{*}{Cyprus Hellenistic}
& DEM                        & \textbf{0.799} & $-$1.19 \\
& Slope                      & \textbf{0.736} & $-$0.75 \\
& Aspect                     & 0.533 & \phantom{$-$}0.13 \\
& Curvature                  & 0.506 & $-$0.01 \\
& Water dist.                & 0.549 & \phantom{$-$}0.19 \\
\bottomrule
\end{tabularx}
\end{table}

Three findings emerge. First, Sagalassos DEM-derived features are near-chance discriminators (AUC 0.50--0.72); the landscape-level terrain does not distinguish Late Antique sites from background. The discriminative signal comes entirely from the \textbf{historical infrastructure features}: distance to Hellenistic roads (AUC $= 0.842$) and the Late Antique settlement distance map (AUC $= 0.790$). These are spatially fine-grained features, namely proximity to specific linear road corridors and known settlement zones, that provide precise pseudo-label targets.

Second, and counterintuitively, \textbf{Cyprus has stronger raw DEM separability than Sagalassos} (DEM AUC $= 0.799$, Slope AUC $= 0.736$ vs.\ Sagalassos DEM AUC $= 0.540$). Hellenistic sites cluster in coastal lowlands (site mean elevation 176\,m vs.\ background 455\,m, $d = -1.19$). However, this signal reflects a \textbf{broad geographic gradient} (coastal lowlands vs.\ mountain interior) rather than fine-grained site-specific signatures. DPL exploits this gradient but cannot distinguish individual site locations from surrounding coastal terrain, causing positive pseudo-labels to spread across the entire lowland zone. This is consistent with the probability surfaces (Figure~\ref{fig:supp:cyprus_uniform_surfaces}), which are dominated by coastal proximity gradients.

Third, Cyprus has \textbf{no historical infrastructure features}: no road networks, no period-specific settlement maps. The absence of spatially specific auxiliary features leaves DPL relying exclusively on coarse terrain gradients, which are insufficient for stable pseudo-label bootstrapping.

\subsection{LAMAP Hyperparameters and Sensitivity}
\label{supp:lamap_params}

LAMAP selects the $K=15$ spatially nearest training sites for each prediction cell and weights their contributions by $w_i = \exp(-d_i / d_{\max} \cdot \lambda)$, where $d_i$ is the distance to site $i$, $d_{\max}$ is the distance to the 15th (furthest selected) site, and $\lambda$ is the decay parameter. We set $\lambda = 1.0$, providing exactly one e-folding of decay across the full 15-site neighbourhood. For the Cyprus Hellenistic dataset, the median nearest-neighbour distance between sites is 0.48\,km and the median distance to the 15th nearest site is 3.0\,km. With $\lambda = 1.0$, the nearest site (median 0.48\,km) receives weight $\approx 0.85$, while the 15th site (3.0\,km) receives weight $\approx 0.37$. The decay is scale-adaptive: $d_{\max}$ adjusts automatically to local site density.

These hyperparameters were not tuned for Cyprus and represent principled defaults carried over from the Sagalassos application. Domain calibration by an archaeologist with knowledge of the Hellenistic settlement system could yield improved performance; in particular, $K$ and $\lambda$ are most sensitive to local settlement patterns. The reported LAMAP results should therefore be interpreted as a lower bound on what a domain-calibrated LAMAP could achieve.

\subsection{Spatial K-Fold Results}
\label{supp:spatial_kfold}

Table~\ref{tab:phase1_cyprus_spatial} reports Cyprus Hellenistic results under spatial k-fold with 4\,km exclusion buffers. All methods show high fold-level variance. SL achieves ROC-AUC $= 0.607 \pm 0.133$ (range 0.311--0.882), nominally exceeding DPL ($0.520 \pm 0.149$, range 0.198--0.776); we attribute this to fold-selection effects rather than consistent learning. SL\_Pos collapses to Recall $= 1.000$ with ROC-AUC $= 0.576 \pm 0.129$.

\begin{table}[h]
\centering
\caption{Cyprus Hellenistic spatial k-fold with 4\,km buffer zones ($r=1$\,m). DL: mean $\pm$ std over 5 folds $\times$ 5 seeds. LAMAP: mean $\pm$ std over 5 folds, 1 seed (deterministic). Bold: best per metric. $^*$\,positive collapse. $^\ddagger$\,LAMAP hyperparameters not tuned for Cyprus.}
\label{tab:phase1_cyprus_spatial}
\begin{tabularx}{\textwidth}{l*{2}{>{\centering\arraybackslash}X}}
\toprule
Model & ROC-AUC & Recall \\
\midrule
SL
& $0.607 \pm 0.133$
& $0.003 \pm 0.008$ \\

SL\_Pos
& $0.576 \pm 0.129$
& $^*1.000 \pm 0.000$ \\

DPL
& $0.520 \pm 0.149$
& $0.239 \pm 0.333$ \\

\textbf{LAMAP}$^\ddagger$
& $\mathbf{0.678 \pm 0.162}$
& $\mathbf{0.679 \pm 0.111}$ \\
\bottomrule
\end{tabularx}
\end{table}

%% file: references.bib
@misc{lirias3727771,
  author = {Willett, Patrick},
  title = {Transforming Landscapes of Southwest Anatolia: Modeling Social and Environmental Change from the Middle to Late Holocene Using Predictive Land-use and Cropland Reconstructions},
  year = {2022}
}

@inbook{c76eb591-d9c7-39f9-8e35-a83c9cadd6a6,
  author = {Jeroen Poblome},
  booktitle = {Documenting Ancient Sagalassos: A Guide to Archaeological Methods and Concepts},
  pages = {7--30},
  publisher = {Leuven University Press},
  title = {Introducing the Sagalassos Archaeological Research Project},
  year = {2023}
}

@inbook{79e093c11d7f4cfa82304dd46566057e,
  title = {The countryside - Where are the people?},
  author = {Ralf Vandam and Eva Kaptijn and Rinse Willet and Patrick Willett},
  year = {2019},
  pages = {262--270},
  booktitle = {Meanwhile in the Mountains: Sagalassos},
  publisher = {Yapı Kredi Yayınları}
}

@inproceedings{elkan2008learning,
  title={Learning classifiers from only positive and unlabeled data},
  author={Elkan, Charles and Noto, Keith},
  booktitle={Proc. 14th ACM SIGKDD Int. Conf. Knowl. Discov. Data Mining},
  pages={213--220},
  year={2008}
}

@article{Bekker_2020,
  title={Learning from positive and unlabeled data: a survey},
  volume={109},
  number={4},
  journal={Mach. Learn.},
  author={Bekker, Jessa and Davis, Jesse},
  year={2020},
  pages={719–760}
}

@article{DBLP:journals/corr/KiryoNPS17,
  author = {Ryuichi Kiryo and Gang Niu and Marthinus Christoffel du Plessis and Masashi Sugiyama},
  title = {Positive-Unlabeled Learning with Non-Negative Risk Estimator},
  journal = {CoRR},
  volume = {abs/1703.00593},
  year = {2017}
}

@article{10.1371/journal.pone.0239424,
  author = {Yaworsky, Peter M. and Vernon, Kenneth B. and Spangler, Jerry D. and Brewer, Simon C. and Codding, Brian F.},
  journal = {PLOS ONE},
  number = {10},
  pages = {1-22},
  title = {Advancing predictive modeling in archaeology: An evaluation of regression and machine learning methods on the Grand Staircase-Escalante National Monument},
  volume = {15},
  year = {2020}
}

@article{pseudolabels,
  author = {Lee, Dong-Hyun},
  year = {2013},
  title = {Pseudo-Label : The Simple and Efficient Semi-Supervised Learning Method for Deep Neural Networks},
  journal = {ICML 2013 Workshop : Challenges in Representation Learning (WREPL)}
}

@misc{luo2022scribblesupervisedmedicalimagesegmentation,
  title={Scribble-Supervised Medical Image Segmentation via Dual-Branch Network and Dynamically Mixed Pseudo Labels Supervision},
  author={Xiangde Luo and Minhao Hu and Wenjun Liao and Shuwei Zhai and Tao Song and Guotai Wang and Shaoting Zhang},
  year={2022}
}

@misc{ronneberger2015unetconvolutionalnetworksbiomedical,
  title={U-Net: Convolutional Networks for Biomedical Image Segmentation},
  author={Olaf Ronneberger and Philipp Fischer and Thomas Brox},
  year={2015}
}

@article{MASEK2020111968,
  author = {Jeffrey G. Masek and Michael A. Wulder and Brian Markham and Joel McCorkel and Christopher J. Crawford and James Storey and Del T. Jenstrom},
  journal = {Remote Sens. Environ.},
  pages = {111968},
  title = {Landsat 9: Empowering open science and applications through continuity},
  volume = {248},
  year = {2020}
}

@article{CARLETON20123371,
  author = {W. Chris Carleton and James Conolly and Gyles Ianonne},
  journal = {J. Archaeol. Sci.},
  number = {11},
  pages = {3371-3385},
  title = {A locally-adaptive model of archaeological potential (LAMAP)},
  volume = {39},
  year = {2012}
}

@article{Stewart_TorchGeo_Deep_Learning_2024,
  author = {Stewart, Adam J. and Robinson, Caleb and Corley, Isaac A. and Ortiz, Anthony and Lavista Ferres, Juan M. and Banerjee, Arindam},
  journal = {ACM Trans. Spat. Algorithms Syst.},
  title = {{TorchGeo}: Deep Learning With Geospatial Data},
  year = {2024}
}

@book{banningArchaeologicalSurvey2002,
  title = {Archaeological {{Survey}}},
  author = {Banning, E. B.},
  year = {2002},
  publisher = {Springer Science \& Business Media}
}

@article{wachtelPredictiveModelingArchaeological2018,
  title = {Predictive Modeling for Archaeological Site Locations: {{Comparing}} Logistic Regression and Maximal Entropy in North {{Israel}} and North-East {{China}}},
  author = {Wachtel, Ido and Zidon, Royi and Garti, Shimon and {Shelach-Lavi}, Gideon},
  year = {2018},
  journal = {J. Archaeol. Sci.},
  volume = {92},
  pages = {28--36}
}

@article{vandamMarginalLandscapesHuman2019,
  title = {?Marginal? {{Landscapes}}: {{Human Activity}}, {{Vulnerability}}, and {{Resilience}} in the {{Western Taurus Mountains}} ({{Southwest Turkey}})},
  author = {Vandam, Ralf and Kaptijn, Eva and Broothaerts, Nils and De Cupere, Bea and Marinova, Elena and Van Loo, Maarten and Verstraeten, Gert and Poblome, Jeroen},
  year = {2019},
  journal = {J. East. Mediterr. Archaeol. Heritage Stud.},
  volume = {7},
  number = {4},
  pages = {432--450}
}

@article{bulawkaDeepLearningbasedDetection2024,
  title = {Deep Learning-Based Detection of Qanat Underground Water Distribution Systems Using {{HEXAGON}} Spy Satellite Imagery},
  author = {Bu{\l}awka, Nazarij and Orengo, Hector A. and {Berganzo-Besga}, Iban},
  year = {2024},
  journal = {J. Archaeol. Sci.},
  volume = {171},
  pages = {106053}
}

@article{landauerArchaeologicalSiteDetection2025,
  title={Archaeological site detection: latest results from a deep learning based Europe wide hillfort search},
  author={Landauer, J{\"u}rgen and Maddison, Simon and Fontana, Giacomo and Posluschny, Axel G},
  journal={J. Comput. Appl. Archaeol.},
  volume={8},
  number={1},
  year={2025}
}

@article{banasiakSemanticSegmentationUNet2022,
  title = {Semantic {{Segmentation}} ({{U-Net}}) of {{Archaeological Features}} in {{Airborne Laser Scanning}}---{{Example}} of the {{Bia{\l}owie{\.z}a Forest}}},
  author = {Banasiak, Pawe{\l} Zbigniew and Berezowski, Piotr Leszek and Zap{\l}ata, Rafa{\l} and Mielcarek, Mi{\l}osz and Duraj, Konrad and Stere{\'n}czak, Krzysztof},
  year = {2022},
  journal = {Remote Sens.},
  volume = {14},
  number = {4},
  pages = {995}
}

@article{rondeauDoesLocallyAdaptiveModel2022,
  title = {Does the {{Locally-Adaptive Model}} of {{Archaeological Potential}} ({{LAMAP}}) Work for Hunter-Gatherer Sites? {{A}} Test Using Data from the {{Tanana Valley}}, {{Alaska}}},
  author = {Rondeau, Rob and Carleton, W. Christopher and Collard, Mark and Driver, Jonathan},
  year = {2022},
  journal = {PLOS ONE},
  volume = {17},
  number = {3}
}

@article{xuSemiSupervisedContrastiveLearning2023,
  title = {Semi-{{Supervised Contrastive Learning}} for {{Remote Sensing}}: {{Identifying Ancient Urbanization}} in the {{South-Central Andes}}},
  author = {Xu, Jiachen and Guo, Junlin and {Zimmer-Dauphinee}, James and Liu, Quan and Shi, Yuxuan and Asad, Zuhayr and Wilkes, D. {\relax Mitchell}. and VanValkenburgh, Parker and Wernke, Steven A. and Huo, Yuankai},
  year = {2023},
  journal = {Int. J. Remote Sens.},
  volume = {44},
  number = {6},
  pages = {1922--1938}
}

@article{ultralyticsYOLOv8,
  title={YOLOv8: The Ultimate YOLO Model},
  author={Ultralytics},
  journal={arXiv preprint arXiv:2304.00501},
  year={2023}
}

@misc{NASA_ASTER_GDEM_V3,
  author = {{NASA/JPL/ASTER}},
  title = {ASTER Global Digital Elevation Model V003},
  year = {2025}
}

@misc{zhao2022distpupositiveunlabeledlearninglabel,
  title={Dist-PU: Positive-Unlabeled Learning from a Label Distribution Perspective},
  author={Yunrui Zhao and Qianqian Xu and Yangbangyan Jiang and Peisong Wen and Qingming Huang},
  year={2022}
}

@article{phillips2009sample,
  title={Sample selection bias and presence-only distribution models: implications for background and pseudo-absence data},
  author={Phillips, Steven J and Dud{\'\i}k, Miroslav and Elith, Jane and Graham, Catherine H and Lehmann, Anthony and Leathwick, John and Ferrier, Simon},
  journal={Ecol. Appl.},
  volume={19},
  number={1},
  pages={181--197},
  year={2009}
}

@inproceedings{Zhao_2023,
  title={Class Prior-Free Positive-Unlabeled Learning with Taylor Variational Loss for Hyperspectral Remote Sensing Imagery},
  booktitle={Proc. IEEE/CVF Int. Conf. Comput. Vis.},
  author={Zhao, Hengwei and Wang, Xinyu and Li, Jingtao and Zhong, Yanfei},
  year={2023},
  pages={16781–16790}
}

@article{crawford2022cyprus,
  title={Cyprus Dataset: Settlements from 11000 BCE to 1878 CE},
  author={Crawford, Katherine A and Vella, Marc-Antoine},
  journal={J. Open Archaeol. Data},
  volume={10},
  year={2022}
}

@article{zhang2025interpretable,
  author = {Yang, Jia and Luo, Lei and Zhao, Jianghong and Ji, Dechang and Sun, Jisi and Fan, Jinhui and Fu, Xingjian and Tu, Ran and Wang, Xinyuan},
  journal = {npj Herit. Sci.},
  number = {1},
  pages = {689},
  title = {Explainable artificial intelligence with negative sample optimization for archaeological site prediction in Surkhandarya Uzbekistan},
  volume = {13},
  year = {2025}
}

@article{10.1093/europace/euad205,
  author = {Six, Stefaan and Theuns, Peter and Libin, Pieter and Now{\'e}, Ann and Pannone, Luigi and Bogaerts, Bart and Jaxy, Simon and Olsen, Catharina and Pappaert, Gudrun and Grau, Isel and Sieira, Juan and Van Dooren, Sonia and Scheirlynck, Esther and Nekkebroeck, Julie and Mallefroy, Marina and de Asmundis, Carlo and Bilsen, Johan},
  title = {Patient-reported outcome measures on mental health and psychosocial factors in patients with Brugada syndrome},
  journal = {EP Europace},
  volume = {25},
  number = {9},
  pages = {euad205},
  year = {2023}
}
